\documentclass[sn-mathphys-num]{sn-jnl}

\usepackage{graphicx}%
\usepackage{lmodern}
\usepackage{anyfontsize}
\usepackage{multirow}%
\usepackage{amsmath,amssymb,amsfonts}%
\usepackage{amsthm}%
\usepackage{mathrsfs}%
\usepackage[title]{appendix}%
\usepackage{xcolor}%
\usepackage{textcomp}%
\usepackage{booktabs}%
\usepackage{algorithm}%
\usepackage{algpseudocode}%
\usepackage{lscape}

\title[From Low Field to High Value]{From Low Field to High Value: Robust Cortical Mapping from Low-Field MRI}

\author*[1]{Karthik Gopinath}\email{kgopinath@mgh.harvard.edu}
\author[1]{Annabel Sorby-Adams}
\author[1]{Jonathan W. Ramirez}
\author[1]{Dina Zemlyanker}
\author[1]{Jennifer Guo}
\author[3]{David Hunt}
\author[3]{Christine L. Mac Donald}
\author[3]{C. Dirk Keene}
\author[5]{Timothy Coalson}
\author[5]{Matthew F. Glasser}
\author[5]{David Van Essen}
\author[1]{Matthew S. Rosen}
\author[1,4]{Oula Puonti}
\author[1]{W. Taylor Kimberly}
\author[1,2,6]{Juan Eugenio Iglesias}

\affil[1]{\orgdiv{Massachusetts General Hospital, Harvard Medical School, United States}}
\affil[2]{\orgdiv{Massachusetts Institute of Technology, United States}}
\affil[3]{\orgdiv{University of Washington, United States}}
\affil[4]{Danish Research Centre for Magnetic Resonance, Copenhagen University Hospital, Denmark}
    \affil[5]{Washington University, United States}
\affil[6]{University College London, United Kingdom\textbf{}}

\begin{document}




\abstract{Three-dimensional reconstruction of cortical surfaces from MRI for subsequent morphometric analysis is fundamental for understanding brain structure. While high-field Magnetic Resonance Imaging (HF-MRI) is the standard in research and clinical settings, its relatively limited availability hinders widespread use. Low-field MRI (LF-MRI), particularly portable systems, offers a cost-effective and accessible alternative. However, existing cortical surface analysis tools, such as FreeSurfer, are optimized for high-resolution HF-MRI and struggle with the lower signal-to-noise ratio (SNR) and resolution of LF-MRI. In this work, we present a machine learning method for 3D reconstruction and analysis of portable LF-MRI scans over a range of contrasts and resolutions. Our method works ``out of the box'' and does not require retraining. It leverages a 3D U-Net trained on synthetic LF-MRI data to predict signed distance functions of the cortical surfaces, followed by geometric processing to ensure topologically accurate reconstructions. We evaluate our approach using paired HF/LF-MRI scans of the same subjects, demonstrating that the accuracy of LF-MRI surface reconstruction depends strongly on acquisition parameters, including contrast type (T1 vs T2), orientation (axial vs isotropic), and resolution. A 3~mm isotropic T2-weighted scan acquired in under 4 minutes, which is comparable in duration to typical HF-MRI acquisitions, yields strong agreement with HF-derived surfaces: surface area correlates at $r=0.96$, cortical parcellations reach a Dice coefficient of $0.98$, and gray matter volume achieves $r=0.93$. Cortical thickness remains more challenging but achieves correlations up to $r=0.70$, reflecting the difficulties of achieving sub-mm precision with $\sim$3$\times$3$\times$3~mm voxels. Our results also show that recon-any performs robustly across other sequences and contrasts, though thickness estimates are particularly sensitive and degrade substantially with anisotropic or low-resolution scans.  We also validate our method on challenging postmortem LF-MRI scans, further illustrating its robustness. Our method represents a significant step toward making cortical surface analysis feasible for portable LF-MRI systems. The tool is publicly available at \url{https://surfer.nmr.mgh.harvard.edu/fswiki/ReconAny}. }


\maketitle

\section{Introduction}
    
\begin{figure}[t]
        \centering
        \includegraphics[width=\textwidth]{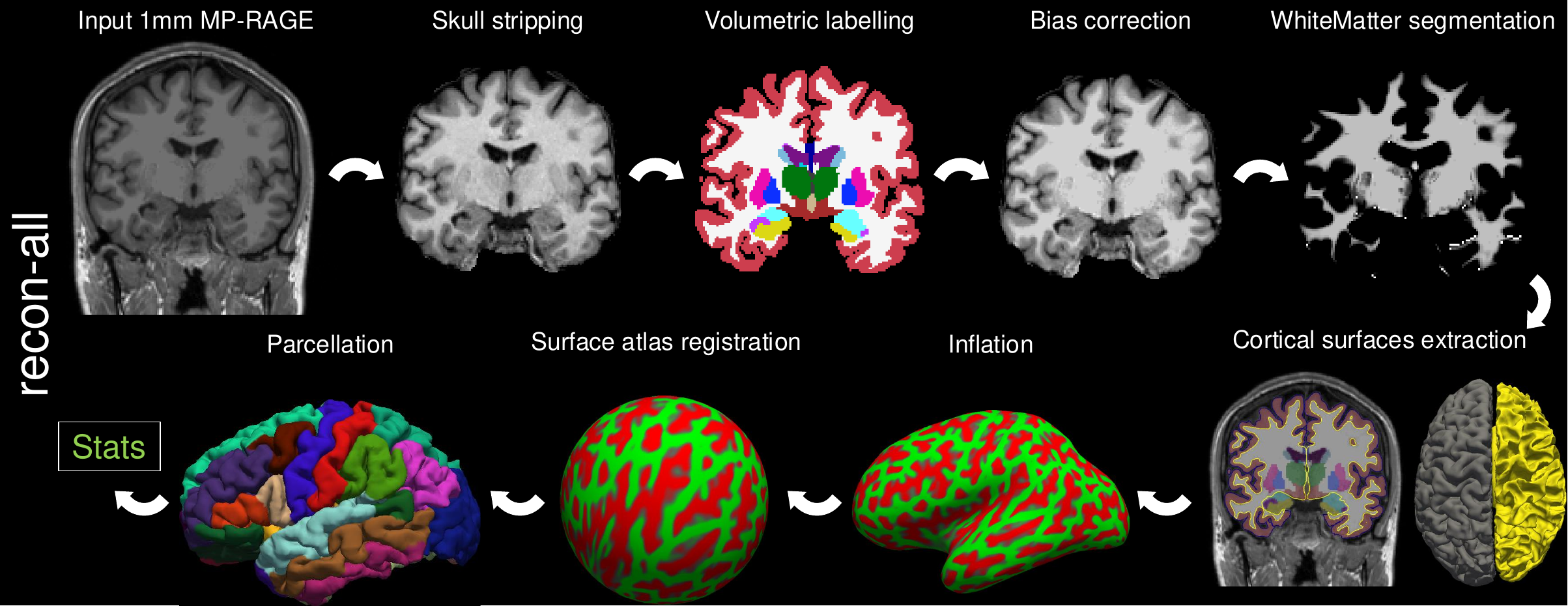}
        \\
        \includegraphics[width=\textwidth]{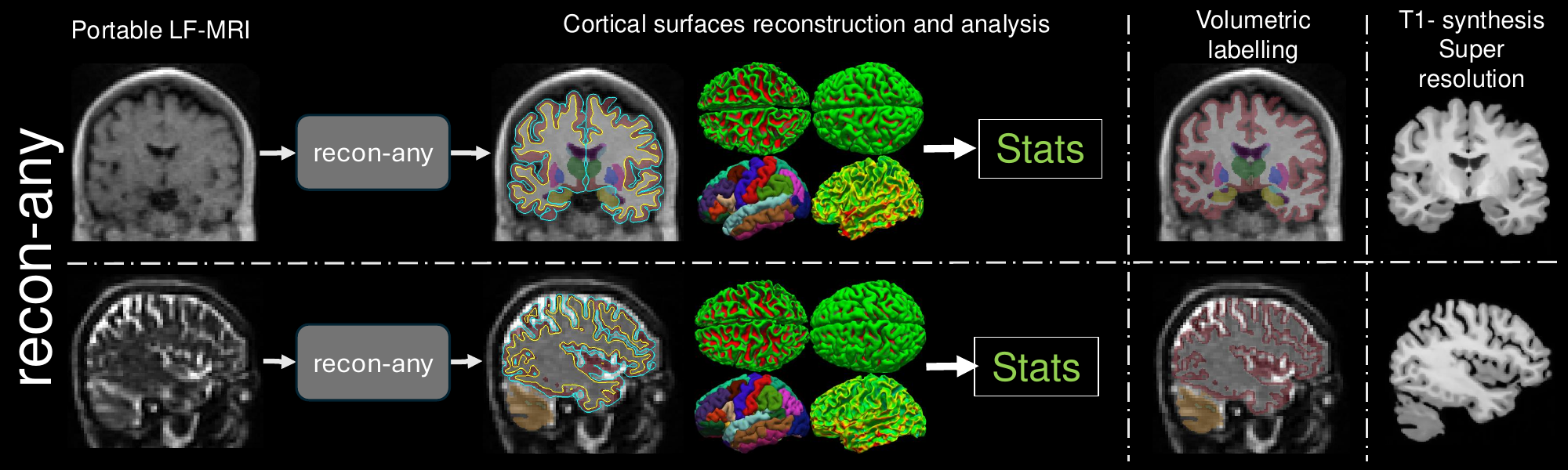}
        \caption{Comparison of a classical cortical surface reconstruction (FreeSurfer's \textit{recon-all}, top) and our deep learning-based approach (Recon-Any, bottom). \textit{Recon-all} is optimized for T1 HF-MRI and follows a multi-step volumetric segmentation and surface extraction process. Recon-Any directly predicts cortical surfaces from LF-MRI via signed distance functions (SDFs), enabling accurate surface reconstruction across different MRI contrasts and resolutions. In addition, recon-any also predicts volumetric segmentation and generates T1 contrast at 1~mm resolution.}
        \label{fig:classical_vs_dl}
    \end{figure}

    The human cerebral cortex is a complex and highly folded structure that plays a fundamental role in cognition, sensory processing, and motor function~\cite{fuster2005cortex,shipp2007structure}. Cortical morphometry, which involves analyzing features such as cortical thickness, surface area, and curvature, provides insightful information on neurodevelopment, aging, and various neurological disorders, including Alzheimer's disease~\cite{salat2004thinning, querbes2009early, rosas2002regional}. These structural biomarkers are widely used in neuroimaging studies to track disease progression and assess treatment efficacy~\cite{eskildsen2015structural}. High-field magnetic resonance imaging (HF-MRI), typically operating at 1.5 Tesla (T) or higher, has long been the gold standard for cortical analysis due to its high spatial resolution and strong contrast between gray matter (GM) and white matter (WM). T1-weighted (T1) imaging, in particular, is well suited for cortical morphometry, as it provides a sharp delineation of the GM-WM boundary. Automated pipelines such as FreeSurfer's \textit{recon-all}~\cite{dale1999cortical, fischl1999cortical}, ANTs~\cite{tustison2014large}, and BrainSuite~\cite{shattuck2002brainsuite} have been developed for cortical processing of HF-MRI data. These follow a structured workflow that generally includes skull stripping, bias field correction, intensity normalization, white matter segmentation, and surface reconstruction, followed by inflation, spherical registration, and cortical parcellation (Figure~\ref{fig:classical_vs_dl}, top). These methods are highly optimized for isotropic, high-resolution MRI scans with T1 contrast, making them reliable tools for neuroimaging research. However, their reliance on HF-MRI infrastructure limits their applicability in settings where advanced imaging infrastructure is not available.

    Recent advances in low-field MRI (LF-MRI) have opened new possibilities for more accessible neuroimaging~\cite{lim_low-field_2024}. Certain portable LF-MRI scanners, operating at field strengths between 10~mT and 100~mT, offer a cost-effective alternative to high-field MRI (HF-MRI) and have demonstrated feasibility for bedside and point-of-care neuroimaging~\cite{sheth_bedside_2022,yuen2022portable}. Some of these systems utilize lightweight, permanent magnet designs, which eliminate the need for cryogenic cooling and complex shielding~\cite{arnold_low-field_2023}. These characteristics have made such devices particularly useful in intensive care units, emergency medicine, and remote or resource-limited environments, where conventional MRI access is limited. The Hyperfine 64~mT scanner exemplifies this category and has been deployed for bedside imaging in acute care contexts, showing promise in assessing neurological conditions such as stroke~\cite{laso2024quantifying,shay2024portable}, traumatic brain injury~\cite{zabinska2024low}, and neurodegenerative diseases~\cite{sorby2024portable}.

    In addition to \emph{in vivo} applications, postmortem brain imaging with portable LF-MRI has emerged as a promising approach in MRI - neuropathology correlation studies. The portability and cost efficiency of portable LF-MRI make it particularly well suited for autopsy settings, where timely imaging is crucial. Unlike high-field MRI scanners, which are often unavailable for postmortem use due to rigid scheduling and continuous clinical demand, portable LF-MRI scanners can be placed within the autopsy suite, enabling rapid on-site imaging. Rapid imaging is important because metabolic degradation of brain tissue begins shortly after death, constraining the window for acquiring usable MRI data. However, only very few select sites have a dedicated high-field MRI scanner for cadaveric imaging. Therefore, LF-MRI has great potential in bridging the gap between \emph{in vivo} MR imaging and histological validation in neurodegenerative disease research. 
    

    Despite these advantages, cortical surface reconstruction from LF-MRI presents considerable technical challenges. The inherently lower magnetic field strength results in a reduced signal-to-noise ratio (SNR), which limits constrast-to-noise ratio and spatial resolution, thus leading to stronger partial volume effects than at high field. This makes it difficult to accurately position the WM and pial surfaces. As in HF-MRI, surface placement in LF-MRI may also be complicated by the fact that, for certain types of MRI contrast (e.g., T1), the  dura mater and GM can exhibit similar signal intensities, leading to further errors in pial surface extraction. As a result, traditional surface reconstruction pipelines such as FreeSurfer's \textit{recon-all} do not produce reliable cortical surface representations from LF-MRI data.
    
    In the context of HF-MRI, deep learning methods have been proposed as an alternative to classical cortical surface extraction pipelines~\cite{henschel2022fastsurfervinn, bongratz2022vox2cortex}. These approaches can be broadly categorized into explicit and implicit surface estimation techniques. Implicit methods produce volumetric outputs from which cortical surfaces are subsequently derived using traditional tessellation-based reconstruction. For example, FastSurfer~\cite{henschel2022fastsurfervinn} employs convolutional neural networks (CNNs) to generate segmentation maps, which are then used to reconstruct surfaces. Other implicit methods, such as DeepCSR~\cite{cruz2021deepcsr} and recon-all-clinical~\cite{gopinath2023cortical}, predict signed distance functions (SDFs), enabling surface extraction via iso-surfacing algorithms. In contrast, explicit methods generate cortical surfaces directly, typically by deforming a template mesh to align with an MRI scan. This strategy avoids the need to resolve complex topological issues during training and accelerates processing. Examples include Vox2Cortex~\cite{bongratz2022vox2cortex}, CortexODE~\cite{ma2022cortexode}, and TopoFit~\cite{hoopes2022topofit}, all of which rely on graph-based neural networks for template deformation. While these deep learning-based methods have proven effective on HF-MRI, they are not designed to handle the resolution and contrast limitations characteristic of LF-MRI.

    Efforts to enable automated analysis of lower resolution MRI scans have followed two main strategies. One approach involves using deep learning super-resolution methods to synthesize high-resolution, T1-like images from LF-MRI scans. These enhanced images can then be processed with standard pipelines such as \textit{recon-all}. Examples of these methods include SynthSR~\cite{iglesias2023synthsr}, which was designed for clinical MRI scans with large slice spacing, or its LF-MRI variant  LF-SynthSR~\cite{iglesias2022quantitative}. While these approaches yield good segmentation accuracy for subcortical brain regions~\cite{sorby2024portable}, they remain suboptimal for cortical surface reconstruction, as they cannot accurately recover sharp cortical geometry. An alternative approach is to directly estimate cortical surfaces from low-resolution scans without an intermediate upsampling step. This strategy relies on learning objectives that directly penalize geometric errors in the predicted surfaces, such as deviations in signed distance functions, rather than voxelwise intensity differences that may not consistently correlate with inaccuracies in surface placement. In our previous work, we introduced \textit{recon-all-clinical}~\cite{gopinath2023cortical}, a method that enables cortical surface reconstruction on clinical-grade MRI scans. This approach demonstrated robustness across variations in resolution contrast and orientation, outperforming traditional segmentation-based methods. However, while \textit{recon-all-clinical} was designed for clinical HF-MRI, it does not fully address the challenges associated with portable LF-MRI.
    
    In this work, we introduce \textit{recon-any}, a novel deep learning-based framework that builds upon the foundation of \textit{recon-all-clinical} to enable robust cortical surface reconstruction, segmentation, and morphometric analysis from any LF-MRI scan. Unlike conventional cortical pipelines optimized for high-field 1~mm T1-weighted MRI, \textit{recon-any} is trained to generalize across various LF-MRI acquisitions, handling arbitrary MRI contrasts (e.g., T1, T2, FLAIR, etc), orientations (sagittal, axial, coronal), resolutions, and scanner-specific artifacts. This is achieved through domain-randomized synthetic training, incorporating \emph{a priori} knowledge of the LF-MRI image formation process, including contrast variations, SNR limitations, and resolution constraints.
    
    As shown in Figure~\ref{fig:classical_vs_dl}, \textit{recon-any} is a comprehensive pipeline that also includes segmentation, T1 contrast synthesis, and super-resolution, making it compatible with FreeSurfer tools. Unlike traditional methods that require explicit volumetric segmentation before surface extraction, \textit{recon-any} directly predicts SDFs to reconstruct cortical surfaces, enabling robust morphometric analysis even at low resolution and SNR.  
    
    We validate our approach using paired HF/LF-MRI scans, demonstrating strong agreement in cortical surface measurements across different contrasts and resolutions. Additionally, we present qualitative results on \emph{postmortem brain imaging} of fresh and cadeveric brain scans acquired with the Hyperfine MRI scanner, highlighting the versatility of our approach. \textit{Recon-any} supports precise cortical surface reconstruction, parcellation, and morphometric analysis in portable LF-MRI, across both \emph{in vivo} and postmortem brain imaging contexts. This represents a step toward making cortical morphometry feasible in resource-limited environments and point-of-care application. The tool is publicly available as part of FreeSurfer and can be accessed at \url{https://surfer.nmr.mgh.harvard.edu/fswiki/ReconAny}.

\section{Results}
    \begin{figure}[t!]
        \centering
        \includegraphics[width=0.991\textwidth]{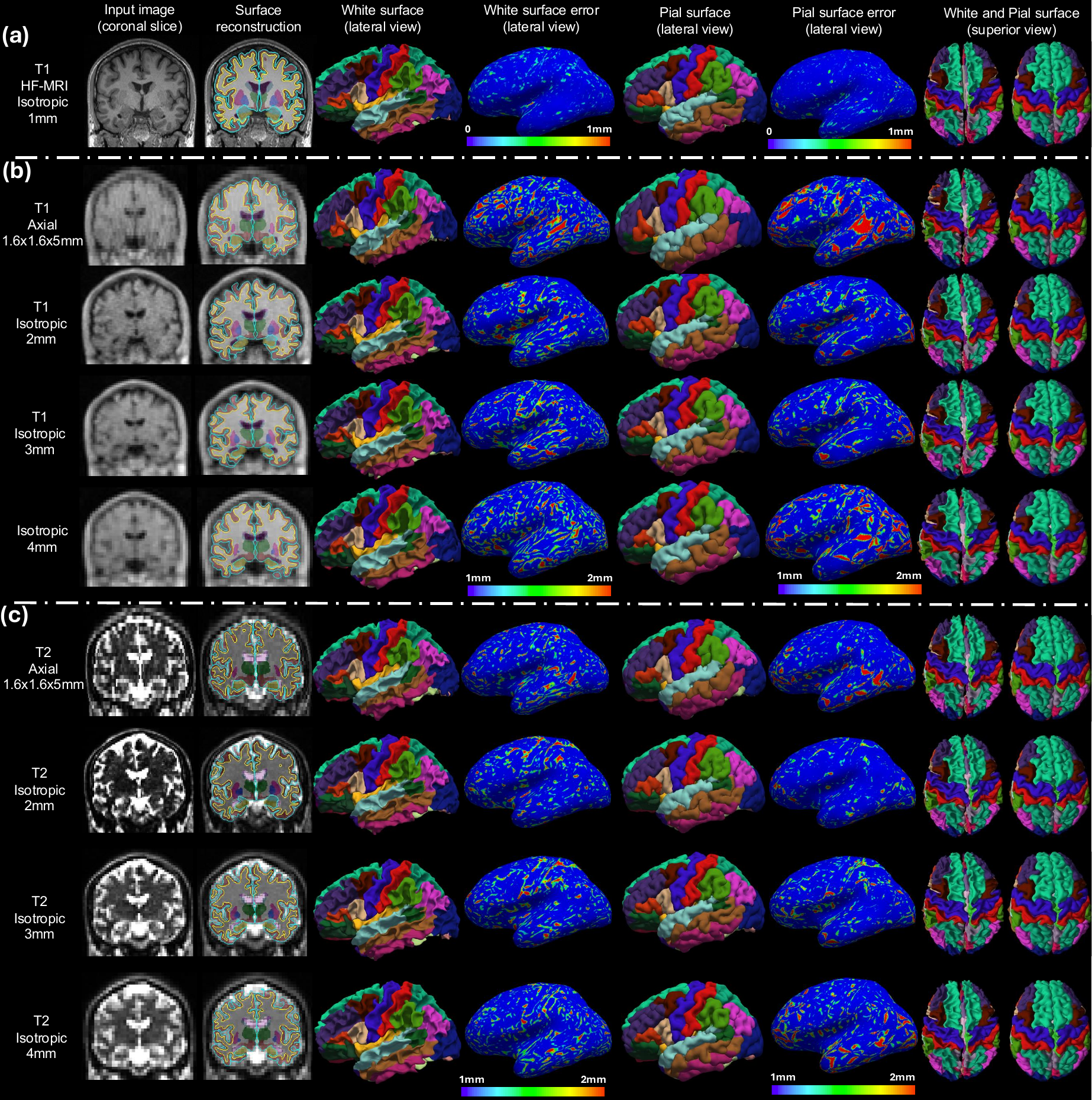}
        \includegraphics[width=0.995\textwidth]{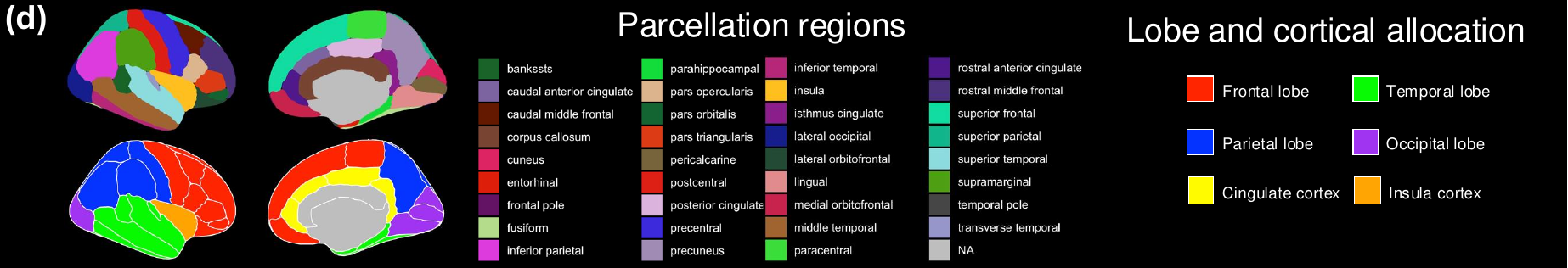}
        \caption{\textbf{Cortical surface reconstruction from LF- and HF-MRI for a sample subject} (a)~HF T1w MRI (1~mm isotropic) processed with FreeSurfer \textit{recon-all} serves as the reference surfaces used for all comparisons. (b)~LF-MRI T1-weighted scans of varying resolutions (1.6~mm~$\times$~1.6~mm~$\times$~5~mm axial, 2~mm, 3~mm, and 4~mm) processed using \textit{recon-any} show strong alignment with HF-derived surfaces, with error maps highlighting minor discrepancies primarily in deep sulci. (c)~T2-weighted LF-MRI scans demonstrate similarly robust reconstruction performance, particularly at 2~mm and 3~mm isotropic resolutions. (d)~Parcellation regions are defined by the Desikan-Killiany atlas~\citep{desikan2006automated}, grouped into anatomical lobes (frontal, parietal, occipital, temporal, cingulate, and insular) for morphometric analysis.}

        \label{fig:qualitative_results}
                
    \end{figure}
    
    We evaluate \textit{recon-any} as a complete cortical analysis pipeline by validating its performance across multiple LF-MRI resolutions and contrasts. Specifically, we assess: (1)~cortical surface reconstruction accuracy by comparing LF-MRI-derived surfaces to those obtained from HF-MRI, (2)~parcellation reliability compared to anatomical labels from HF-MRI surfaces, and (3)~the agreement of regional morphometric measurements including gray matter volume, surface area, and cortical thickness between LF-MRI and HF-MRI derived reconstructions. As a reference for evaluating the performance of \textit{recon-any} on lower-resolution portable LF-MRI data, we use cortical surfaces reconstructed from 1~mm isotropic T1 HF-MRI scans processed with FreeSurfer's \textit{recon-all} pipeline.

    \subsection{Evaluation of Cortical Surface Reconstruction}
    

        Figure~\ref{fig:qualitative_results} shows a qualitative comparison of cortical surface reconstructions between HF-MRI and LF-MRI scans, which illustrates the ability of \textit{recon-any} to accurately delineate cortical surfaces across different resolutions and contrasts. 

        To quantitatively evaluate the accuracy of cortical surface reconstruction, we use 15 high-resolution 1~mm isotropic T1-weighted HF-MRI scans and paired LF-MRI scans (including 1.6~mm~$\times$~1.6~mm~$\times$~5~mm axial, 2~mm, 3~mm, and 4~mm T1 and T2 scans) from the same subjects (60 scans total). We compare LF-MRI surfaces reconstructed with \textit{recon-any} to those obtained from HF-MRI using FreeSurfer's~\citep{fischl1999cortical} \textit{recon-all} pipeline. We quantify surface reconstruction accuracy using absolute average distance (AAD) and 90th percentile Hausdorff distance (HD90) for both the WM and pial surfaces.  
         \begin{figure}[t!]
            \centering
            \includegraphics[width=\textwidth]{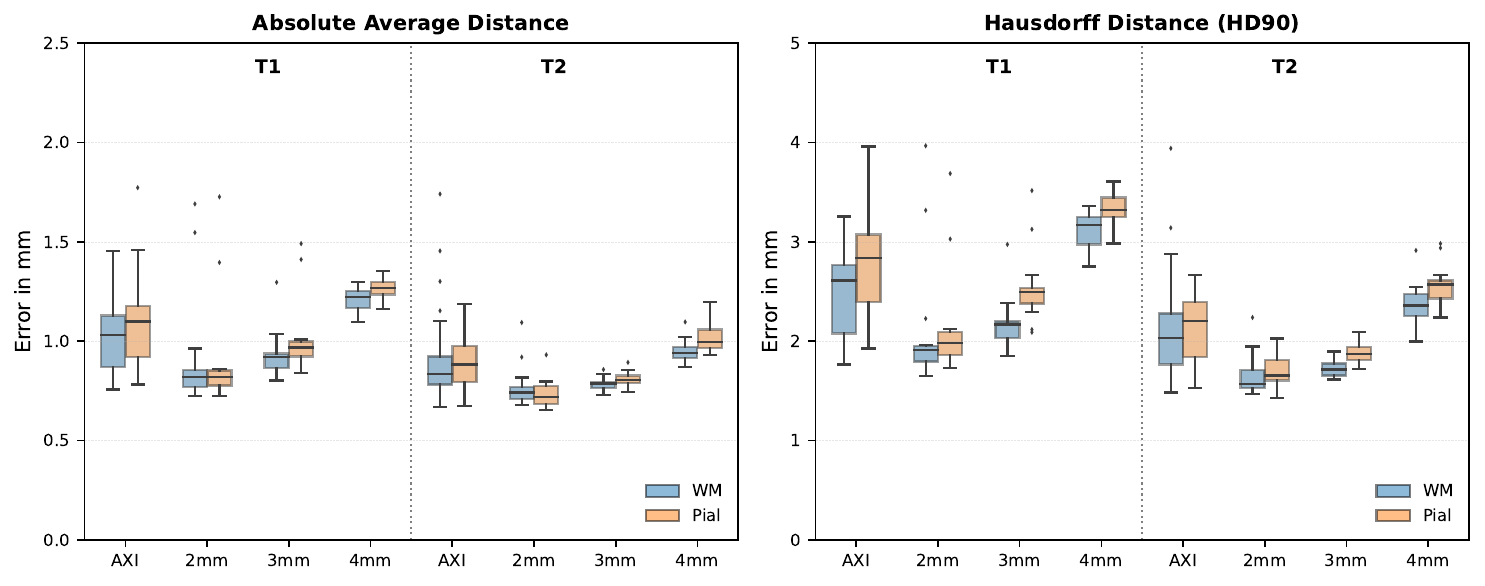}
            \caption{Cortical surface reconstruction errors across different LF-MRI resolutions and contrasts. Left: Absolute Average Distance (AAD). Right: 90th percentile Hausdorff Distance (HD90). Results are shown for both white matter (WM, blue) and pial (orange) surfaces across low-field T1 and T2 MRI scans with 1.6~mm~$\times$~1.6~mm~$\times$~5~mm axial, 2~mm, 3~mm, and 4~mm isotropic voxel sizes. Each box indicates the interquartile range, with the median shown as a central horizontal line. The results demonstrate that T2 scans consistently show lower errors than their T1 scans. For reference, the best-case surface reconstruction accuracy on 1~mm isotropic HF-MRI (used during training) yielded mean AAD values of 0.365~mm (white) and 0.395~mm (pial), and HD90 values of 0.709~mm and 0.852~mm, respectively.}
            \label{fig:surface_accuracy}
        \end{figure}

        Despite the lower resolution and contrast of LF-MRI, the reconstructed white and pial surfaces closely match those derived from HF-MRI, maintaining consistent cortical geometry. The error maps in the Figure~\ref{fig:qualitative_results} highlight minor deviations, predominantly in deep sulcal regions, where partial volume effects and tissue contrast limitations are most pronounced. The error maps also highlight that the default axial sequences of the Hyperfine scanner are suboptimal for cortical analysis, compared with isotropic acquisitions available via its advanced model.

        Figure~\ref{fig:surface_accuracy} presents the distribution of reconstruction errors across different LF-MRI voxel sizes (1.6~mm~$\times$~1.6~mm~$\times$~5~mm axial, 2mm, 3mm, 4mm) for both T1 and T2 scans. Remarkably, the average AAD errors of \textit{Recon-any} range from 1~mm to 2~mm, which is well below the voxel dimensions at LF-MRI. When using isotropic resolution, the error decreases with voxel size, with diminishing returns below 3~mm. Notably, while the 2~mm isotropic scan requires a much longer acquisition time (approximately 12 minutes compared to 4 minutes for the 3~mm scan), it does not yield proportionally improved surface accuracy. As also observed in Figure~\ref{fig:qualitative_results}, the default axial sequences yield inferior performance in cortical surface placement.  Figure~\ref{fig:surface_accuracy} also shows that the T2 sequences yield better surface estimation than their T1 counterparts. These results are consistent with those reported in~\cite{sorby2024portable} in the context of subcortical segmentation, and showcase our method's ability to provide reliable cortical reconstructions in settings where HF-MRI is unavailable.
        
       To contextualize these findings, we also evaluated \textit{recon-any} on the 15 high-resolution HF-MRI scans and compared its outputs to FreeSurfer's \textit{recon-all}, establishing an upper bound of achievable accuracy. In this setting, the mean AAD errors were 0.365~mm for the white surface and 0.395~mm for the pial surface, with HD90 values of 0.709~mm (white) and 0.852~mm (pial). These results reflect the best-case performance of our method and serve as a benchmark for assessing generalization to LF-MRI, and are indicated in Figure~\ref{fig:surface_accuracy} for comparison. Notably, both T1- and T2-weighted scans yield similarly surface reconstruction accuracy across resolutions indicating the robustness of \textit{recon-any} to contrast variability.


    \subsection{Cortical Surface Parcellation}

\begin{figure}[t!]
            \centering
            \includegraphics[width=\textwidth]{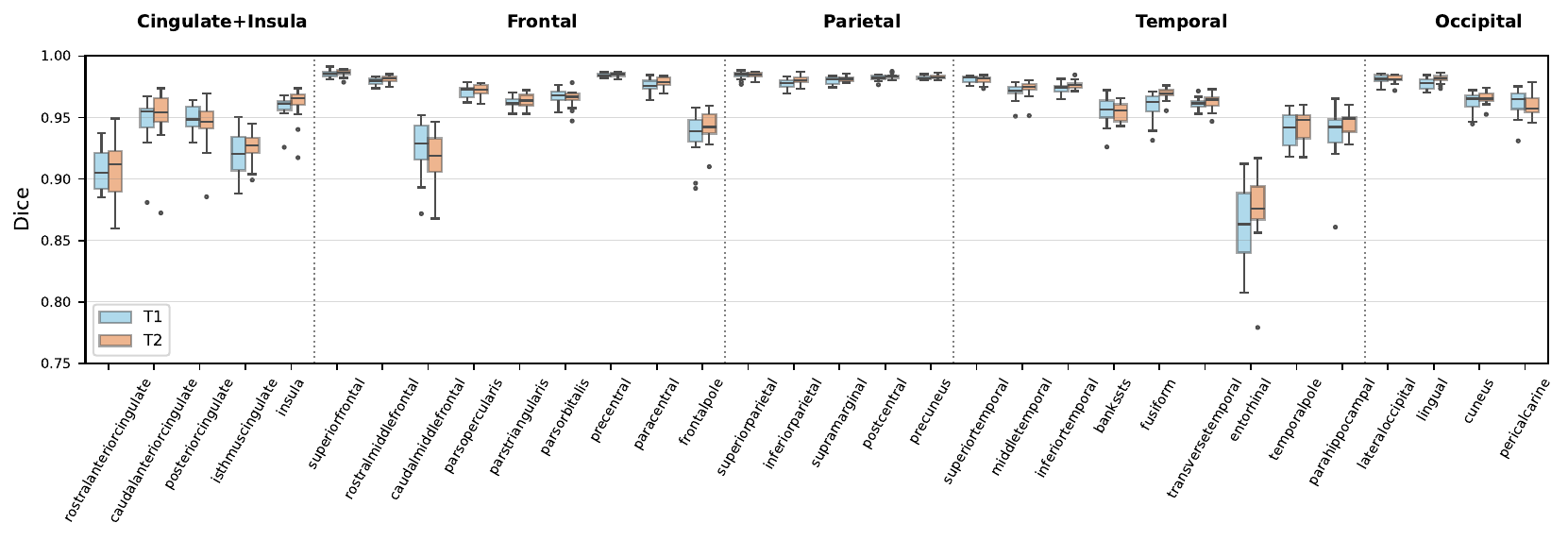}
            \caption{\textbf{Cortical parcellation accuracy.} Dice similarity coefficients (DSC) between cortical parcellations derived from LF-MRI and the reference parcellations from 1~mm isotropic T1 scans processed with FreeSurfer. Results are grouped by anatomical lobe (Cingulate+Insula, Frontal, Parietal, Temporal, Occipital) and shown separately for T1 (blue) and T2 (orange) LF-MRI scans. Each box plot summarizes the distribution of DSCs across 15 subjects: the boxes represent the interquartile range, the horizontal line within each box denotes the median, and individual points beyond this range are shown as outliers. The results highlight consistently high overlap (DSC~$>$~0.85) across most cortical regions, with slightly lower performance in smaller or entorhinal parcel.}
            \label{fig:parcellation_accuracy}
        \end{figure}

        Cortical parcellation accuracy was evaluated by computing the Dice similarity coefficients (DSC) between cortical regions segmented from LF-MRI and HF-MRI. The results in this section are based on 3~mm isotropic T1 and T2 LF-MRI scans, which offer a favorable balance between acquisition time (under 5 minutes) and reconstruction accuracy. To facilitate analysis, cortical regions were grouped into anatomical lobes, including the frontal, temporal, parietal, occipital, and insula/cingulate cortex. The mean Dice coefficient was computed for each parcel across all resolutions to assess the reliability of parcellation. Figure~\ref{fig:parcellation_accuracy} shows the distribution of Dice coefficients for different cortical lobes, while results for other LF-MRI resolutions are provided in Appendix~\ref{fig:cor_parc_app1} and Appendix~\ref{fig:cor_parc_app2}.

        Our method, \textit{recon-any}, demonstrates high spatial overlap between LF-MRI and HF-MRI-derived parcellations, with Dice coefficients consistently exceeding 0.85 for all major cortical structures. The strongest agreement was observed in the parietal and occipital lobes across different MRI contrasts and resolutions. The frontal and temporal lobes also showed high consistency, with Dice values above 0.90 in most cases. T1 and T2 scans exhibited similar performance across all lobes, indicating that \textit{recon-any} generalizes well across contrast types.
          
        Smaller cortical structures, such as the entorhinal cortex, exhibited relatively lower Dice scores (0.87) -- but still high in absolute terms. The entorhinal cortex is among the most challenging regions to segment due to its complex folding patterns, relatively small volume, and lower contrast at reduced field strengths. These factors contribute to increased variability in its segmentation, even in HF-MRI. Despite this, \textit{recon-any} maintains high segmentation accuracy across diverse imaging conditions, highlighting its outstanding robustness for automated parcellation in portable LF-MRI settings.

    \subsection{Automated Morphometric Quantification}  

\begin{figure}[t!]
            \centering
            \includegraphics[width=\textwidth]{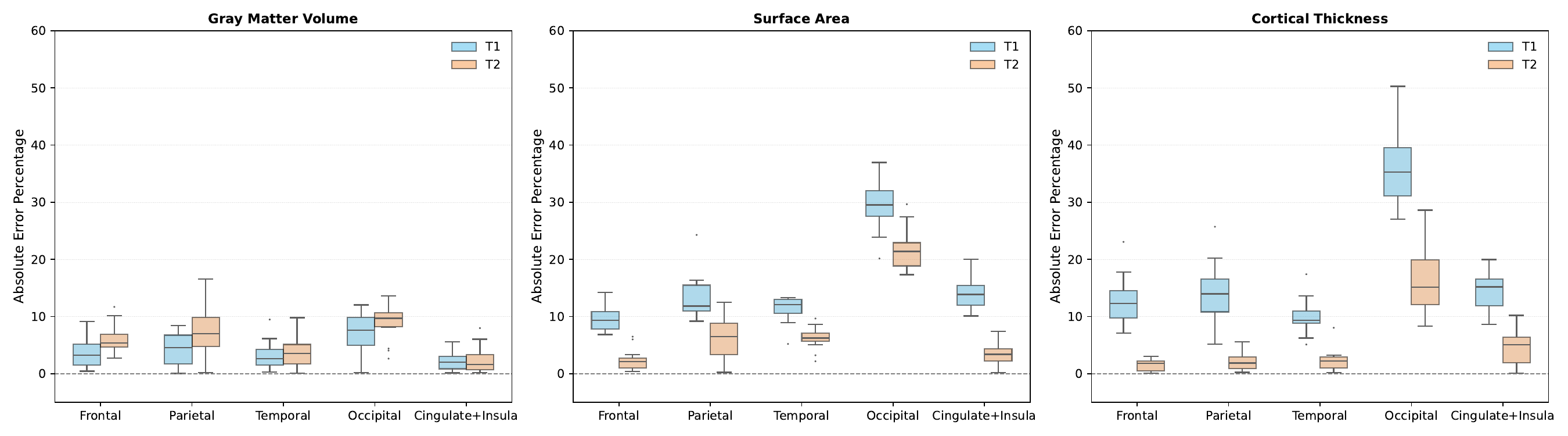}
            \caption{\textbf{Cortical morphometry accuracy.} Absolute percentage error between morphometric estimates from LF-MRI and the HF-MRI gold standard across major cortical lobes. Three metrics are shown: gray matter volume (left), surface area (middle), and cortical thickness (right), with separate box plots for T1 (blue) and T2 (orange) scans acquired at 3~mm isotropic resolution. Each box plot summarizes the distribution across 15 subjects: boxes represent the interquartile range and points beyond that are plotted as outliers. Surface area and volume estimates show strong agreement (typically within 10\%), while cortical thickness shows higher variability, particularly in the occipital lobe, reflecting the sensitivity of thickness estimation to resolution and contrast.}

            \label{fig:morphometry_accuracy}
        \end{figure}

        Cortical morphometric measures, including gray matter volume, cortical surface area, and cortical thickness, were estimated using \textit{recon-any} and compared against gold standard values derived from HF-MRI. 
        Figure~\ref{fig:morphometry_accuracy} presents errors in surface area, gray matter volume, and cortical thickness for different lobes, computed from 3~mm isotropic LF-MRI scans for both T1- and T2-weighted acquisitions. Table~\ref{tab:morphometry_correlations} displays the corresponding Pearson correlation coefficients between the same morphometric measures, derived from LF-MRI and HF-MRI at different resolutions. Both the figure and the table indicate that gray matter volume and surface area can be accurately estimated using \textit{recon-any}. Results for other LF-MRI resolutions are provided in Appendix~\ref{fig:cort_morpho_allres}. The average absolute error in gray matter volume remains below 10\% across all lobes, with typical values such as 3--6\% in the frontal and temporal lobes. Surface area estimates show slightly larger deviations, generally within 10--25\%, with the largest errors observed in the occipital lobe. The corresponding Pearson correlations exceed 0.90 across all lobes and imaging contrasts and often exceed 0.95.  
        
        Cortical thickness estimation, on the other hand, yields considerably larger errors and lower correlations. The absolute error percentages range from 15--30\% in most lobes and can reach up to 50\% in the occipital lobe for certain T1 scans. This increased variability is expected, given that the typical cortical thickness ranges between 2--4~mm, and detecting sub-millimeter variations with LF-MRI is inherently challenging due to resolution limitations. Despite this, the thickness estimates produced by \textit{recon-any} follow expected anatomical patterns and yield meaningful correlations, particularly for isotropic LF T2 scans (r~$>$~0.7). Correlations for T1 LF-MRI are weaker (r~$\sim$~0.3), but remain statistically significant. The strongest agreement is observed in the frontal and temporal lobes, while greater variability occurs in the cingulate and insular cortices---regions that are prone to inter-subject variability and MRI-related artifacts. Overall, T2-weighted scans consistently yield lower error rates than their T1 counterparts across all lobes and metrics, likely due to their superior gray–CSF contrast, as seen in Figure~\ref{fig:qualitative_results}. These results further highlight the robustness of \textit{recon-any} in recovering reliable surface-based morphometry from diverse LF-MRI acquisitions.

        
        \begin{table}[ht!]
            \centering
            \setlength{\tabcolsep}{5pt}  
            \footnotesize  
            \caption{Pearson correlation coefficients for gray matter volume, surface area, and cortical thickness across different LF-MRI resolutions. Each resolution (1.6$\times$1.6$\times$5.0~mm axial (AXI), 2~mm, 3~mm, 4~mm) is evaluated separately for T1 and T2 scans. Higher correlations indicate stronger agreement between LF-MRI and HF-MRI measurements. Values in brackets denote 95\% confidence intervals. $^\dagger$ denotes non-significant correlations.}
            \label{tab:morphometry_correlations}
            \begin{tabular}{@{}ccccccc@{}}
            \toprule
            \multirow{2}{*}{Resolution} & \multicolumn{2}{c}{Gray Matter Volume} & \multicolumn{2}{c}{Surface Area} & \multicolumn{2}{c}{Cortical Thickness} \\
            \cmidrule(lr){2-3} \cmidrule(lr){4-5} \cmidrule(lr){6-7}
            & T1 & T2 & T1 & T2 & T1 & T2 \\
            \midrule
            AXI   & 0.94[.82–.98] & 0.95[.85–.98] & 0.95[.85–.98] & 0.93[.80–.98] & 0.34[-.21–.73]$^\dagger$ & 0.42[-.12–.77]$^\dagger$ \\
            2$mm$ & 0.94[.82–.98] & 0.97[.91–.99] & 0.97[.91–.99] & 0.98[.94–.99] & 0.32[-.23–.72]$^\dagger$ & 0.61[.14–.86] \\
            3$mm$ & 0.95[.85–.98] & 0.95[.85–.98] & 0.95[.85–.98] & 0.96[.88–.99] & 0.30[-.25–.70]$^\dagger$ & 0.70[.29–.89] \\
            4$mm$ & 0.94[.82–.98] & 0.92[.77–.97] & 0.92[.77–.97] & 0.90[.72–.97] & 0.33[-.22–.72]$^\dagger$ & 0.63[.17–.86] \\
            \bottomrule
            \end{tabular}
        \end{table}

    \subsection{Postmortem LF-MRI Surface Reconstruction}

\begin{figure}[t!]
                \centering
                \includegraphics[width=\textwidth]{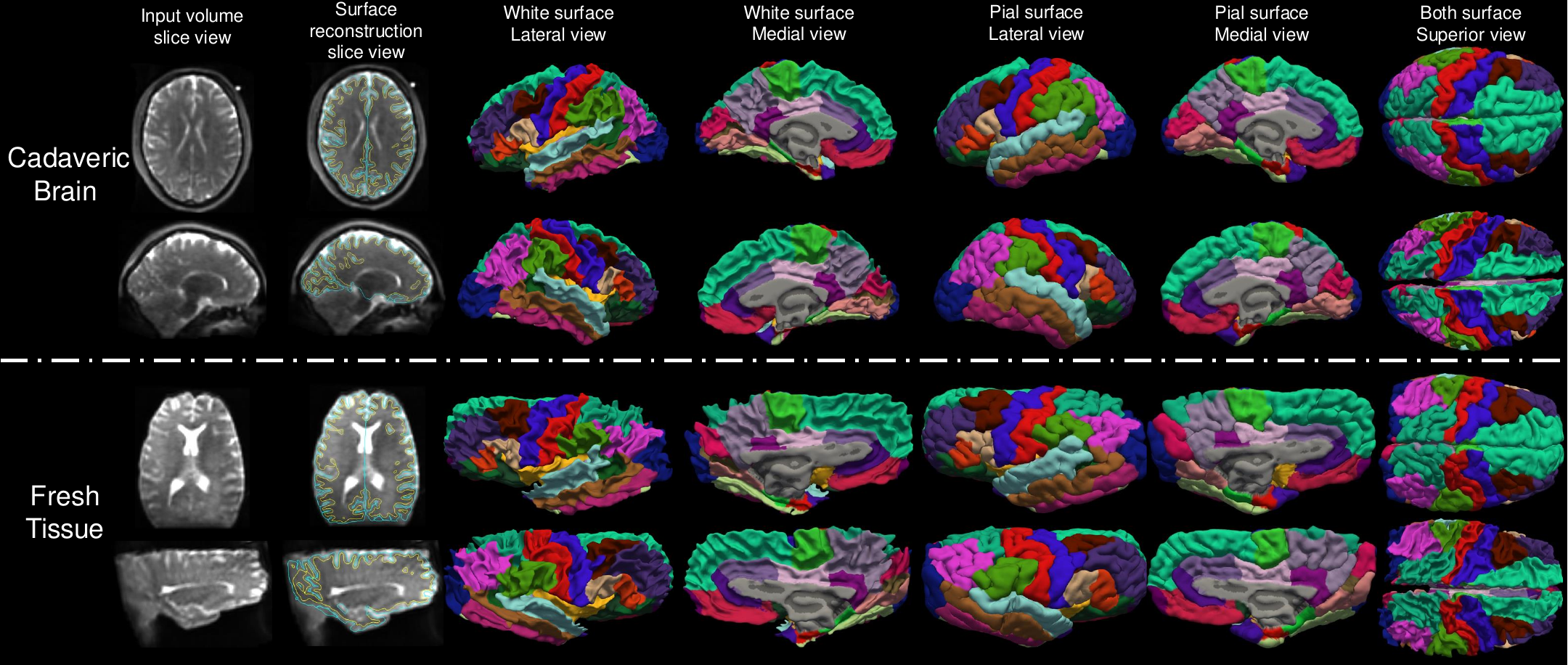}
                    \caption{\textbf{Postmortem cortical reconstruction.} LF-MRI scans of cadaveric and fresh brain tissue acquired at 2.3~mm isotropic resolution and processed with \textit{recon-any} yield accurate cortical surfaces and parcellations. The method preserves sulcal and gyral geometry even in fresh tissue and generates accurate reconstructions despite heavy deformation and signal loss in the occipital region.}
                \label{fig:qualitative_ex_vivo_results}
            \end{figure}

        \textit{Recon-any} enables automated cortical surface analysis from portable LF-MRI and extends its applicability to postmortem neuroimaging, encompassing both cadaveric and freshly extracted brains. Cadaveric brains retain structural integrity due to the presence of intracranial support, minimizing deformation. In contrast, fresh brains scanned after extraction exhibit more pronounced collapse and shape distortion from the lack of cerebrospinal fluid and physical containment. Despite these challenges, Figure~\ref{fig:qualitative_ex_vivo_results} demonstrates that \textit{recon-any} successfully reconstructs cortical surfaces across both conditions. Sulcal and gyral patterns are preserved even in the greatly deformed fresh brain in the figure, and surface parcellations remain anatomically plausible. 
        
        Direct validation against HF-MRI is impractical in postmortem settings, as it would require access to high-field scanners within the autopsy suite and rapid imaging workflows, both of which are often unavailable, as discussed in the introduction. As a result, we rely on quality control provided by expert neuroanatomists to assess the anatomical plausibility of the cortical reconstructions and parcellations. We applied \textit{recon-any} to a cohort of 31 postmortem LF-MRI scans, comprising both cadaveric and fresh brains with varying degrees of structural deformation. Some fresh brains exhibited substantial shape distortion due to the absence of intracranial support, as illustrated in Figure~\ref{fig:qualitative_ex_vivo_results}. Despite these challenges, all 31 cases passed visual quality control. Two representative examples, one a cadaveric brain and a second challenging deformed fresh brain are shown in Figure~\ref{fig:qualitative_ex_vivo_results}. The reconstructions for the 29 remaining cases are presented in the Supplementary Materials.
        
        These results demonstrate that \textit{recon-any} enables high-fidelity cortical surface reconstruction in postmortem portable LF-MRI, thereby extending surface-based morphometric analysis to scenarios where traditional in vivo imaging is impossible. This facilitates anatomical assessment and supports potential integration with histological or biochemical postmortem analyses.
    
\section{Discussion}\label{sec12}

    Cortical surface reconstruction and morphometric analysis have traditionally relied on HF-MRI, where standardized, high-resolution scans enable precise delineation of cortical geometry. While recent deep learning methods have accelerated these workflows, they remain limited to HF-MRI and assume specific MRI contrasts, typically T1. Our findings demonstrate that \textit{recon-any} successfully extends cortical surface analysis to LF-MRI, overcoming the challenges posed by reduced resolution, lower SNR, and contrast variability. By integrating \emph{a priori} knowledge of the LF-MRI image formation process during training, \textit{recon-any} reconstructs cortical surfaces with high fidelity across a wide spectrum of LF-MRI acquisitions.
    
    A key outcome of this study is the robust agreement between LF-MRI and HF-MRI-derived cortical measurements, despite the inherent limitations of LF imaging. Surface reconstruction errors remained within $\sim$1~mm of HF-MRI baselines, indicating that cortical geometry can be reliably extracted even from low-resolution scans. Parcellation accuracy was notably high, with over 30 of 34 regions achieving Dice scores above 0.90, highlighting the model’s capability to segment cortical regions despite reduced tissue contrast. Moreover, gray matter volume and surface area estimates showed strong correlation with HF-MRI, affirming that LF-MRI can yield morphometric measures comparable to those from research-grade imaging. Cortical thickness estimates, while anatomically plausible, showed lower consistency with HF-MRI due to their sensitivity to resolution and contrast limitations in LF-MRI.

    Interestingly, our results indicate that T2-weighted LF-MRI yields more accurate surface reconstructions and stronger morphometric correlations with HF-MRI than does T1-weighted LF-MRI. At high field, T1 MRI is typically preferred due to its strong gray–white matter contrast, which facilitates WM surface delineation. At low field, this contrast is diminished in both T1 and T2 scans (see Figure~\ref{fig:qualitative_results}), and the superior CSF–gray matter contrast in T2-weighted scans becomes more useful for accurate pial surface estimation. These findings suggest that T2-weighted sequences may be preferable for cortical analysis in LF-MRI, a consideration that should guide future acquisition protocols and scanner design.

    Despite these challenges, cortical thickness estimation remains feasible with \textit{recon-any}, though its accuracy is more sensitive to acquisition parameters than other morphometric measures. Thickness correlations with HF-MRI are strongest for isotropic T2-weighted scans ($r=0.70$), where higher gray–CSF contrast improves delineation of the pial surface. In contrast, performance drops for anisotropic T1-weighted acquisitions, where reduced resolution and contrast yield weaker correlations (as low as $r=0.30$). These results underscore the importance of careful acquisition design for accurate thickness estimation at low field.

    It is important to note that all evaluations in this work were conducted on LF-MRI data acquired using Hyperfine scanners. While these devices represent the most widely available portable LF-MRI platform, future work will investigate generalizability to other LF-MRI systems and acquisition protocols. The ability to perform fully automated cortical surface analysis on portable LF-MRI represents a major step toward democratizing neuroimaging. Current tools are incompatible with the resolution and contrast variability of LF-MRI, rendering \textit{recon-any} the only viable method currently available for cortical analysis in these settings. 

\section{Methods} \label{sec:methods}
    \textit{Recon-any} consists of several components: \textit{(i)}~generation of synthetic training data using domain randomization to simulate LF-MRI-like acquisitions; \textit{(ii)}~a deep learning model that predicts signed distance functions (SDFs) for cortical boundaries; \textit{(iii)}~surface reconstruction from predicted SDFs followed by post-processing; and \textit{(iv)}~downstream cortical parcellation and morphometric analysis. This section describes each component in detail, along with the datasets and evaluation metrics used to assess performance.

    \textit{Recon-any} extends the capabilities of \textit{recon-all-clinical}~\cite{gopinath2024recon}, which was originally designed for clinical MRI scans with varying contrasts and resolutions -- particularly anisotropic acquisitions with large slice spacing. While \textit{recon-all-clinical} demonstrated robustness across diverse clinical-grade MRI acquisitions, it was not optimized for the specific challenges posed by LF-MRI: lower signal-to-noise ratio, reduced spatial resolution, and altered tissue contrast compared with higher field scans.
    Like \textit{recon-all-clinical}, \textit{recon-any} also employs a domain-randomized synthetic training framework, but seeks to mimic the resolution, contrast, and noise characteristics typically found in portable LF-MRI acquisitions. This approach enables robust generalization across LF-MRI imaging protocols without requiring acquisition-specific modeling. Additionally, unlike \textit{recon-all-clinical}, which was not developed for postmortem brain imaging, \textit{recon-any} also includes simulations of \emph{postmortem} brains during training. This enables \textit{recon-any} to  robustly analyze cortical properties from not only \emph{in vivo}, but also postmortem LF-MRI scans -- further broadening its utility.

    The recon-any pipeline performs cortical surface analysis, including surface reconstruction, parcellation, and morphometric estimation. Similar to FreeSurfer's \textit{recon-all}, \textit{recon-any} reconstructs white and pial surfaces, and derives cortical registrations and parcellations. In addition, it synthesizes a high-resolution T1-like image, which can be used for other analyses. A key distinction is that, rather than relying on explicit segmentation-based surface extraction, \textit{recon-any} builds upon the SDF-based methodology of \textit{recon-all-clinical}, leveraging a deep learning model trained on synthetic LF-like data. 
    
\begin{figure}[t]
        \centering
        \includegraphics[width=\textwidth]{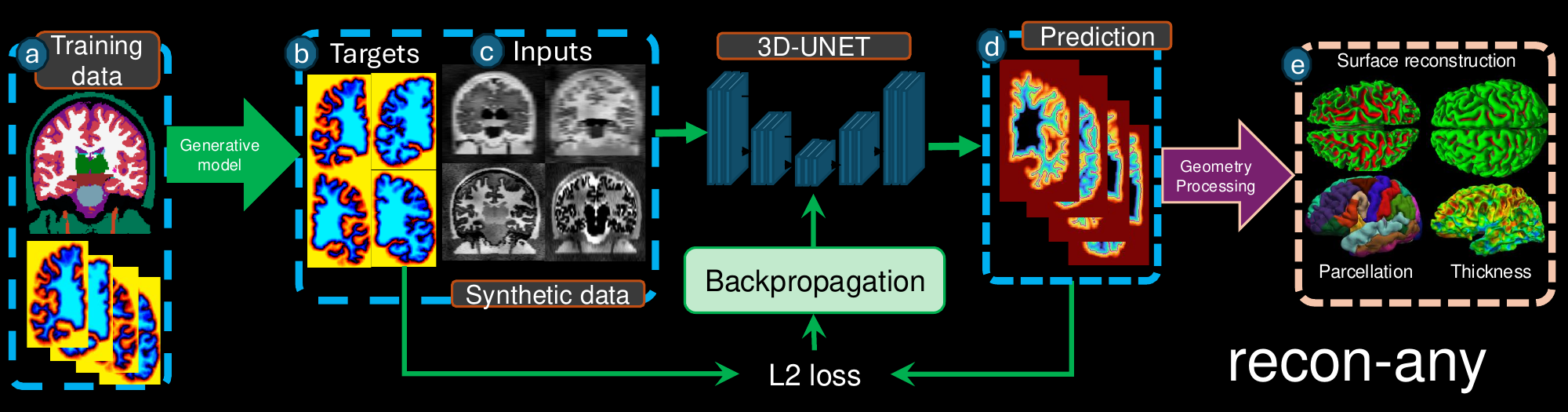}
        \caption{\textbf{Overview of \textit{recon-any}}. The pipeline consists of five main stages labeled (a)–(e). \textbf{(a)} High-resolution brain segmentations and cortical surfaces generated automatically (using FreeSurfer) from high-resolution datasets such as HCP and ADNI are used to build a training dataset. \textbf{(b)} Signed distance functions (SDFs) to the white and pial surfaces are computed from (a); these are used as regression targets during training. \textbf{(c)} Synthetic LF-MRI images are created for each epoch randomly using a generative model that simulates contrast, resolution and noise properties typical of LF-MRI. These synthetic images serve as inputs to the model during training. \textbf{(d)} A 3D U-Net is trained to predict the SDFs from the synthetic LF-like inputs using an $L_2$ loss. Predicted SDFs are then used to reconstruct surfaces. \textbf{(e)} Final outputs include cortical surface reconstructions, parcellations via atlas-based registration, and morphometric measures such as cortical thickness, surface area and gray matter volume.}
        \label{fig:overview}
    \end{figure}

    Figure~\ref{fig:overview} illustrates the full \textit{recon-any} pipeline. 
    A low-field scan, irrespective of contrast or resolution, is processed through a 3D U-Net trained on domain-randomized synthetic data. The network predicts signed distance functions (SDFs), from which white and pial surfaces are extracted using a marching cubes algorithm. A topology correction step ensures smooth and anatomically consistent cortical surfaces. Additionally, the pipeline yields an automated segmentation that can be used to compute volumes of brain regions, as well as a super-resolved 1~mm isotropic T1 scan that can be used for further image analysis, e.g., image registration~\cite{iglesias2023synthsr}. The rest of this section more thoroughly discusses these building blocks, with a focus on the  key extensions that enable robust cortical surface reconstruction in LF-MRI; further details on the original \textit{recon-all-clinical} components can be found in~\cite{gopinath2024recon}.
        
    \subsection{Training Data Generation}
    
        Training data were derived from two high-resolution MRI datasets: the Human Connectome Project (HCP)~\cite{glasser2013minimal} and the Alzheimer's Disease Neuroimaging Initiative (ADNI)~\cite{jack2008alzheimer}. These datasets provide complementary demographic distributions, with HCP comprising predominantly younger, healthy individuals, while ADNI includes older adults, including cognitively normal controls as well as individuals with mild cognitive impairment (MCI) and Alzheimer's disease. Both datasets include high-resolution ($\sim$1~mm isotropic) T1-weighted scans that were used to generate gold standard surfaces and segmentations. We randomly selected 500 subjects from each dataset for training. Specifically, T1-weighted MRI scans were processed using the SAMSEG algorithm~\cite{PUONTI2016235} to generate volumetric segmentations, and FreeSurfer's \textit{recon-all} pipeline (version $7.4.0$) to extract white and pial surfaces. Signed distance functions were computed from these surfaces, encoding the distance of each voxel from the respective cortical boundary. These SDFs served as ground truth labels during model training. We emphasize that the original high-resolution images were not used directly, but rather were used to generate anatomically accurate segmentations and surfaces for subsequent image synthesis.

        The ADNI scans used in the preparation of this article were obtained from the ADNI database (\url{http://adni.loni.usc.edu}). The ADNI was launched in 2003 as a public-private partnership, led by Principal Investigator Michael W. Weiner, MD, with the primary goal of testing whether serial MRI, PET, other biomarkers, and clinical and neuropsychological assessments can be combined to measure the progression of mild cognitive impairment (MCI) and early Alzheimer’s disease (AD). For up-to-date information, see \url{http://www.adni-info.org}. To thoroughly evaluate the robustness of our method, we conducted comprehensive tests across three datasets. 
        
    \subsection{Test Datasets and Evaluation}
    
        \textit{Recon-any} was evaluated on paired HF/LF-MRI scans acquired from 15 healthy adults using a 64mT portable MRI system and a 3T MRI scanner. Healthy volunteers were aged 29 $\pm$ 7 years, including 10 Females and 5 Males, who were either White $(n=12)$ or Asian $(n=2)$, non-Hispanic $(n=11)$, or of unknown ethnicity $(n=3)$. The full LF-MRI sequence parameters and acquisition times, including isotropic and axial T1- and T2-weighted sequences, are provided in Table~\ref{tab:lfmri-params} in the Supplementary Appendix. The cortical surfaces derived from HF-MRI using FreeSurfer's \textit{recon-all} pipeline served as the reference standard for evaluating LF-MRI reconstructions. 
        
        In addition, fresh and cadaveric brain scans were acquired using the same Hyperfine 64mT portable MRI system, allowing for further validation of cortical surface reconstructions under controlled conditions. A total of 32 postmortem brains were analyzed, comprising 21 fresh and 11 cadaveric brain scans. The fresh brains were scanned shortly after extraction in the pathology room, while the cadaveric brains remained intact within the cranium. The specific acquisition parameters for the postmortem T2-weighted imaging protocol are detailed in Table~\ref{tab:lfmri-params} in the Supplementary Appendix.
    
    \subsection{Cortical Surface reconstruction and analysis algorithm}
        
        \noindent \textbf{Synthetic data generation via domain randomization:} To address the variability inherent in real-world LF-MRI data, a domain-randomization strategy was employed to generate a diverse set of synthetic training samples~\cite{gopinath2024synthetic}. This approach ensures robustness to variations in MRI contrast, resolution, and acquisition artifacts. In each training iteration a high-resolution segmentation was selected, which was then subjected to geometric transformations, including affine and nonlinear diffeomorphic deformations. These transformed segmentations were used to synthesize MRI images through a generative model inspired by Bayesian segmentation, in which voxel intensities were sampled from a Gaussian mixture model conditioned on tissue labels, and further combined with artifacts including noise and bias field~\cite{billot2023synthseg,gopinath2024recon}. 
        
        To mimic the acquisition properties of LF-MRI, we made four main modifications with respect to ~\cite{billot2023synthseg,gopinath2024recon}:
        \begin{itemize}
        \item First, we used a specific resolution model. When simulating anisotropic images, the original model never sampled in-plane resolutions coarser than 1~mm. In \textit{recon-any}, we consider in-plane resolutions as low as 2~mm. This strategy effectively models the anisotropic acquisitions typical of LF-MRI devices, e.g., the ``stock'' sequences in the Hyperfine device have  resolution. Furthermore, we consider isotropic resolutions of voxel size up to 4.0~mm -- whereas the original model did not consider sizes beyond 1.5~mm. 
        
        \item Second, we simulated bias fields that can be twice as strong as those simulated in the original model. This strategy endows the trained neural network with robustness against the strong signal drops that are more prominent at low field, e.g., the limited signal sensitivity of the coil in inferior regions in the Hyperfine device.
        
        \item Third, we simulated nonlinear deformations that can be twice as strong as those in ~\cite{billot2023synthseg,gopinath2024recon}. This is crucial when analyzing postmortem cases, particularly fresh brains that may suffer from strong deformation outside of the skull. To better model \emph{ex vivo} cases, we also simulated rotations twice as big compared with the original model, since positioning can vary much more dramatically than when scanning \emph{in vivo}.
        
        \item The fourth and final modification was the simulation of brain images lacking extracerebral tissue, including images without cerebellum or brainstem, and images of single hemispheres. This equips our trained model with the ability to analyze a wide array of postmortem cases.
        \end{itemize}
        
        \noindent \textbf{Deep Learning Model and Training:} We employed a 3D U-Net architecture~\cite{cciccek20163d} for predicting the 1~mm isotropic SDFs from the input scans. The U-Net's encoder-decoder structure is well suited for capturing both local and global features required for accurate surface reconstruction. The architecture consists of four encoding and four decoding levels, incorporating skip connections to preserve spatial information, each comprising two convolutional blocks with $3 \times 3 \times 3$ kernels. The number of feature maps per level increases in the encoder path as $[32, 64, 128, 256]$ and symmetrically decreases in the decoder path. The network was optimized using an L2 loss function, which penalizes the squared differences between the predicted and ground truth SDF values given by $L_2 = \sum_{i \in \mathcal{N}} (\widehat{SDF}_i - SDF_i)^2$, where \( \widehat{SDF}_i \) is the predicted value for voxel \( i \), and \( SDF_i \) is the ground truth value. To focus the training on the relevant boundary regions, SDF values were clipped to a maximum absolute distance of 5~mm from the surface. The model was optimized using the Adam optimizer~\cite{kingma2014adam} with a learning rate of \(10^{-4}\), and training was performed for 300,000 iterations on an NVIDIA Tesla V100 GPU. \\

        \noindent \textbf{Surface Extraction and Analysis:} At test time, SDFs are first predicted from an input scan using the trained U-Net. A first estimate of the white matter surface is obtained with the marching cubes algorithm~\cite{lorensen1998marching}. This estimate is then refined with an algorithm that numerically optimizes an objective function in order to improve the alignment of the surface with the SDF, while encouraging smoothness and preventing self-intersections~\cite{gopinath2024synthetic}. The pial surface is expanded outward from the white matter boundary to the gray-CSF boundary using a similar objective function. Post-processing corrects topological errors, such as handles or holes~\cite{dale1999cortical} and maintains smoothness, preventing self-intersections in the 3D surfaces. Further details can be found in~\cite{gopinath2024synthetic} \\

        \noindent \textbf{Parcellation and Morphometric Estimation:} After extracting the cortical surface, we performed morphometric analyses using FreeSurfer~\citep{fischl2012freesurfer}. Surface area was computed by summing the areas of all triangles in the cortical mesh, while gray matter volume was estimated as the volume enclosed between the white matter and pial surfaces. Cortical thickness was computed by measuring the shortest distance from each point on the white matter surface to the pial surface, and vice versa. The final thickness at each vertex was obtained by averaging these two distances, yielding a symmetric and robust estimate across the cortex. For cortical parcellation, the reconstructed surfaces were registered to the FreeSurfer \textit{fsaverage} template~\citep{fischl2004automatically} in the spherical domain. This enabled segmentation into anatomical regions using the Desikan-Killiany atlas~\citep{desikan2006automated}, providing regional cortical thickness and volume measurements for further analysis. \\

        \noindent \textbf{Evaluation Metrics:} We assessed the accuracy and robustness of cortical surface reconstructions using four different metrics:
        \textbf{Average Surface Distance} (ASD) quantifies the mean Euclidean distance from each point on one surface to its closest point on the other surface. It measures the bidirectional discrepancy between LF-MRI and HF-MRI cortical surfaces and is defined as:
        \begin{equation*}
        \text{ASD}(A, B) = \frac{1}{2} \left( \frac{1}{|A|} \sum_{a \in A} \min_{b \in B} ||a - b|| + \frac{1}{|B|} \sum_{b \in B} \min_{a \in A} ||b - a|| \right)
        \end{equation*}

        \noindent where \( A \) and \( B \) are the sets of vertices from the LF-MRI and HF-MRI cortical surfaces, respectively, and \( ||a - b|| \) denotes the Euclidean distance between points. A lower ASD value indicates a closer alignment between the LF-MRI reconstructed and HF-MRI reference cortical surfaces.

        \textbf{Hausdorff Distance} measures the largest Euclidean distance between the closest points on LF-MRI and HF-MRI surfaces. Since HD is sensitive to outliers, we compute the 90th percentile Hausdorff Distance (HD90), which is a more robust measure of reconstruction error:
        \begin{equation*}
            \text{HD}_{90}(A, B) = P_{90} \left( \max \left\{ \max_{a \in A} \min_{b \in B} ||a - b||, \max_{b \in B} \min_{a \in A} ||b - a|| \right\} \right)
        \end{equation*}        
        \noindent where \( \min_{b \in B} ||a - b|| \) represents the shortest Euclidean distance from a point \( a \) on surface \( A \) to the closest point in \( B \), ensuring each point on \( A \) has a corresponding nearest neighbor on \( B \). Similarly, \( \max_{a \in A} \min_{b \in B} ||a - b|| \) computes the largest of these minimum distances, capturing the worst-case mismatch from \( A \) to \( B \).

        \textbf{Dice Similarity Coefficient} measures the spatial overlap between LF-MRI and HF-MRI-derived surface parcellations obtained from FreeSurfer's \textit{recon-all} pipeline. A DSC of 1 indicates perfect agreement, while a value close to 0 indicates little to no overlap.        
        \begin{equation*}
            \text{DSC} = \frac{2 | A \cap B |}{| A | + | B |}
        \end{equation*}        
        \noindent where \( A \) and \( B \) are the sets of surface nodes with a certain label, estimated from LF-MRI and HF-MRI, respectively. Because these surfaces are reconstructed independently and are not in point-to-point correspondence, we first register both to the same spherical template surface. Dice is then computed on the template, allowing a consistent vertex-wise comparison of labeled regions.

        \textbf{Pearson Correlation Coefficient (r)} evaluates the linear relationship between two variables. We use it to analyze the agreement between surface area and gray matter volume measurements from LF-MRI and HF-MRI. A  value close to r=1 indicates strong agreement, whereas values near r=0 indicate poor correlation.        
        \begin{equation*}
            r = \frac{\sum_{i=1}^{n} (X_i - \overline{X})(Y_i - \overline{Y})}{\sqrt{\sum_{i=1}^{n} (X_i - \overline{X})^2 \sum_{i=1}^{n} (Y_i - \overline{Y})^2}}
        \end{equation*}        
        \noindent where \( X \) and \( Y \) represent the measurements (e.g., surface area or volume for LF-MRI and HF-MRI). 


    \subsection{Data and Code Availability}
        The datasets used in this study for training our \textit{recon-any} model include publicly available high-field MRI scans from the Human Connectome Project (HCP)~\cite{glasser2013minimal} and the Alzheimer's Disease Neuroimaging Initiative (ADNI)~\cite{jack2008alzheimer} datasets. For portable LF-MRI analysis, the paired high-field (3T) and low-field (64mT, Hyperfine) MRI scans used for evaluation in this work were collected under an institutional research protocol and are not publicly available due to participant privacy restrictions and ongoing regulatory constraints. Access to the HCP and ADNI datasets requires approval from their respective data use agreements. Individual patient data can be accessed by academic researchers under restricted conditions, requiring an institutional data use agreement. 
        
        The \textit{recon-any} pipeline is integrated into FreeSurfer and is freely available for research use. The source code, trained models, and documentation are provided at \url{https://surfer.nmr.mgh.harvard.edu/fswiki/ReconAny}. After sourcing FreeSurfer software, \textit{recon-any} can be executed using the following command:  
        \begin{center}
        \texttt{recon-any -i <INPUT\_SCAN> -subjid <SUBJECT\_ID> -side <SIDE> -threads <THREADS> -sd <SUBJECT\_DIR>}
        \end{center}
        
        \noindent where the command-line options are defined as follows:
        \begin{itemize}
            \item \texttt{INPUT\_SCAN}: Path to the MRI image to be processed.
            \item \texttt{SUBJECT\_ID}: Identifier for the subject where a corresponding output directory is created.
            \item \texttt{SIDE}: Specifies the hemisphere(s) to process:
            \begin{itemize}
                \item \texttt{left-c}: Left cerebrum (\emph{postmortem} single hemisphere).
                \item \texttt{left-ccb}: Left cerebrum, cerebellum, and brainstem (if intact).
                \item \texttt{right-c}: Right cerebrum.
                \item \texttt{right-ccb}: Right cerebrum, cerebellum, and brainstem.
                \item \texttt{both}: Both hemispheres (\emph{in vivo} or full \emph{postmortem} brains).
            \end{itemize}
            \item \texttt{THREADS} (optional): Number of CPU threads (default: 1); higher values speed up processing.
            \item \texttt{SUBJECT\_DIR} (optional): Alternative output directory, required if \texttt{SUBJECTS\_DIR} is not set or needs overriding.
        \end{itemize}

\section*{Acknowledgments}
    This work is primarily funded by the National Institute of Aging (1R01AG070988, 1RF1AG080371 and 1R21NS138995). Further support is provided by, BRAIN Initiative (1RF1MH123195, 1UM1MH130981), National Institute of Biomedical Imaging and Bioengineering (1R01EB031114). This work was supported in part by the National Institutes of Health grant 1R21CA267315 (PI Matthew S. Rosen). Matthew S. Rosen acknowledges the generous support of the Kiyomi and Ed Baird MGH Research Scholar award. OP is supported by a grant from the Lundbeck foundation (R360–2021–39). AS-A is supported by the Fulbright Commission and the American Heart Association-Tedy’s Team postdoctoral fellowship. MSR is a founder and equity holder of Hyperfine, Inc. MSR has a financial interest in DeepSpin GmbH. These interests were reviewed and are managed by Massachusetts General Hospital and Mass General Brigham in accordance with their conflict of interest policies.

    Data were provided in part by the Human Connectome Project, WU-Minn Consortium (Principal Investigators: David Van Essen and Kamil Ugurbil; 1U54MH091657) funded by the 16 NIH Institutes and Centers that support the NIH Blueprint for Neuroscience Research; and by the McDonnell Center for Systems Neuroscience at Washington University. Data collection and sharing for this project was also funded by the Alzheimer’s Disease Neuroimaging Initiative (ADNI) (National Institutes of Health Grant U01 AG024904) and DOD ADNI (Department of Defense award number W81XWH-12-2-0012). ADNI is funded by the National Institute on Aging, the National Institute of Biomedical Imaging and Bioengineering, and through generous contributions from the following: AbbVie, Alzheimer’s Association, Alzheimer’s Drug Discovery Foundation, Araclon Biotech, BioClinica, Inc., Biogen, Bristol-Myers Squibb Company, CereSpir, Inc., Cogstate, Eisai Inc., Elan Pharmaceuticals, Inc., Eli Lilly and Company, EuroImmun, F. Hoffmann-La Roche Ltd and its affiliated company Genentech, Inc., Fujirebio, GE Healthcare, IXICO Ltd., Janssen Alzheimer Immunotherapy Research \& Development, LLC., Johnson \& Johnson Pharmaceutical Research \& Development LLC., Lumosity, Lundbeck, Merck \& Co., Inc., Meso Scale Diagnostics, LLC., NeuroRx Research, Neurotrack Technologies, Novartis Pharmaceuticals Corporation, Pfizer Inc., Piramal Imaging, Servier, Takeda Pharmaceutical Company, and Transition Therapeutics. The Canadian Institutes of Health Research provides funding support for ADNI clinical sites in Canada. Private sector contributions are facilitated by the Foundation for the National Institutes of Health (\url{http://www.fnih.org}). The grantee organization is the Northern California Institute for Research and Education, and the study is coordinated by the Alzheimer’s Therapeutic Research Institute at the University of Southern California. ADNI data are disseminated by the Laboratory for Neuro Imaging at the University of Southern California.


\bibliography{References2}


\begin{thebibliography}{38}
\ifx \bisbn   \undefined \def \bisbn  #1{ISBN #1}\fi
\ifx \binits  \undefined \def \binits#1{#1}\fi
\ifx \bauthor  \undefined \def \bauthor#1{#1}\fi
\ifx \batitle  \undefined \def \batitle#1{#1}\fi
\ifx \bjtitle  \undefined \def \bjtitle#1{#1}\fi
\ifx \bvolume  \undefined \def \bvolume#1{\textbf{#1}}\fi
\ifx \byear  \undefined \def \byear#1{#1}\fi
\ifx \bissue  \undefined \def \bissue#1{#1}\fi
\ifx \bfpage  \undefined \def \bfpage#1{#1}\fi
\ifx \blpage  \undefined \def \blpage #1{#1}\fi
\ifx \burl  \undefined \def \burl#1{\textsf{#1}}\fi
\ifx \doiurl  \undefined \def \doiurl#1{\url{https://doi.org/#1}}\fi
\ifx \betal  \undefined \def \betal{\textit{et al.}}\fi
\ifx \binstitute  \undefined \def \binstitute#1{#1}\fi
\ifx \binstitutionaled  \undefined \def \binstitutionaled#1{#1}\fi
\ifx \bctitle  \undefined \def \bctitle#1{#1}\fi
\ifx \beditor  \undefined \def \beditor#1{#1}\fi
\ifx \bpublisher  \undefined \def \bpublisher#1{#1}\fi
\ifx \bbtitle  \undefined \def \bbtitle#1{#1}\fi
\ifx \bedition  \undefined \def \bedition#1{#1}\fi
\ifx \bseriesno  \undefined \def \bseriesno#1{#1}\fi
\ifx \blocation  \undefined \def \blocation#1{#1}\fi
\ifx \bsertitle  \undefined \def \bsertitle#1{#1}\fi
\ifx \bsnm \undefined \def \bsnm#1{#1}\fi
\ifx \bsuffix \undefined \def \bsuffix#1{#1}\fi
\ifx \bparticle \undefined \def \bparticle#1{#1}\fi
\ifx \barticle \undefined \def \barticle#1{#1}\fi
\bibcommenthead
\ifx \bconfdate \undefined \def \bconfdate #1{#1}\fi
\ifx \botherref \undefined \def \botherref #1{#1}\fi
\ifx \url \undefined \def \url#1{\textsf{#1}}\fi
\ifx \bchapter \undefined \def \bchapter#1{#1}\fi
\ifx \bbook \undefined \def \bbook#1{#1}\fi
\ifx \bcomment \undefined \def \bcomment#1{#1}\fi
\ifx \oauthor \undefined \def \oauthor#1{#1}\fi
\ifx \citeauthoryear \undefined \def \citeauthoryear#1{#1}\fi
\ifx \endbibitem  \undefined \def \endbibitem {}\fi
\ifx \bconflocation  \undefined \def \bconflocation#1{#1}\fi
\ifx \arxivurl  \undefined \def \arxivurl#1{\textsf{#1}}\fi
\csname PreBibitemsHook\endcsname

\bibitem[\protect\citeauthoryear{Fuster}{2005}]{fuster2005cortex}
\begin{bbook}
\bauthor{\bsnm{Fuster}, \binits{J.M.}}:
\bbtitle{Cortex and Mind: Unifying Cognition}.
\bpublisher{Oxford university press}, \blocation{???}
(\byear{2005})
\end{bbook}
\endbibitem

\bibitem[\protect\citeauthoryear{Shipp}{2007}]{shipp2007structure}
\begin{barticle}
\bauthor{\bsnm{Shipp}, \binits{S.}}:
\batitle{Structure and function of the cerebral cortex}.
\bjtitle{Current Biology}
\bvolume{17}(\bissue{12}),
\bfpage{443}--\blpage{449}
(\byear{2007})
\end{barticle}
\endbibitem

\bibitem[\protect\citeauthoryear{Salat et~al.}{2004}]{salat2004thinning}
\begin{barticle}
\bauthor{\bsnm{Salat}, \binits{D.H.}},
\bauthor{\bsnm{Buckner}, \binits{R.L.}},
\bauthor{\bsnm{Snyder}, \binits{A.Z.}},
\bauthor{\bsnm{Greve}, \binits{D.N.}},
\bauthor{\bsnm{Desikan}, \binits{R.S.}},
\bauthor{\bsnm{Busa}, \binits{E.}},
\bauthor{\bsnm{Morris}, \binits{J.C.}},
\bauthor{\bsnm{Dale}, \binits{A.M.}},
\bauthor{\bsnm{Fischl}, \binits{B.}}:
\batitle{Thinning of the cerebral cortex in aging}.
\bjtitle{Cerebral cortex}
\bvolume{14}(\bissue{7}),
\bfpage{721}--\blpage{730}
(\byear{2004})
\end{barticle}
\endbibitem

\bibitem[\protect\citeauthoryear{Querbes et~al.}{2009}]{querbes2009early}
\begin{barticle}
\bauthor{\bsnm{Querbes}, \binits{O.}},
\bauthor{\bsnm{Aubry}, \binits{F.}},
\bauthor{\bsnm{Pariente}, \binits{J.}},
\bauthor{\bsnm{Lotterie}, \binits{J.-A.}},
\bauthor{\bsnm{D{\'e}monet}, \binits{J.-F.}}, \betal:
\batitle{Early diagnosis of {Alzheimer}'s disease using cortical thickness: impact of cognitive reserve}.
\bjtitle{Brain}
\bvolume{132}(\bissue{8}),
\bfpage{2036}--\blpage{2047}
(\byear{2009})
\end{barticle}
\endbibitem

\bibitem[\protect\citeauthoryear{Rosas et~al.}{2002}]{rosas2002regional}
\begin{barticle}
\bauthor{\bsnm{Rosas}, \binits{H.}},
\bauthor{\bsnm{Liu}, \binits{A.}},
\bauthor{\bsnm{Hersch}, \binits{S.}},
\bauthor{\bsnm{Glessner}, \binits{M.}},
\bauthor{\bsnm{Ferrante}, \binits{R.}},
\bauthor{\bsnm{Salat}, \binits{D.}},
\bauthor{\bsnm{Der~Kouwe}, \binits{A.}},
\bauthor{\bsnm{Jenkins}, \binits{B.}},
\bauthor{\bsnm{Dale}, \binits{A.}},
\bauthor{\bsnm{Fischl}, \binits{B.}}:
\batitle{Regional and progressive thinning of the cortical ribbon in {Huntington}’s disease}.
\bjtitle{Neurology}
\bvolume{58}(\bissue{5}),
\bfpage{695}--\blpage{701}
(\byear{2002})
\end{barticle}
\endbibitem

\bibitem[\protect\citeauthoryear{Eskildsen et~al.}{2015}]{eskildsen2015structural}
\begin{barticle}
\bauthor{\bsnm{Eskildsen}, \binits{S.F.}},
\bauthor{\bsnm{Coup{\'e}}, \binits{P.}},
\bauthor{\bsnm{Fonov}, \binits{V.S.}},
\bauthor{\bsnm{Pruessner}, \binits{J.C.}},
\bauthor{\bsnm{Collins}, \binits{D.L.}},
\bauthor{\bsnm{Initiative}, \binits{A.D.N.}}, \betal:
\batitle{Structural imaging biomarkers of alzheimer's disease: predicting disease progression}.
\bjtitle{Neurobiology of aging}
\bvolume{36},
\bfpage{23}--\blpage{31}
(\byear{2015})
\end{barticle}
\endbibitem

\bibitem[\protect\citeauthoryear{Dale et~al.}{1999}]{dale1999cortical}
\begin{barticle}
\bauthor{\bsnm{Dale}, \binits{A.M.}},
\bauthor{\bsnm{Fischl}, \binits{B.}},
\bauthor{\bsnm{Sereno}, \binits{M.I.}}:
\batitle{Cortical surface-based analysis: {I}. segmentation and surface reconstruction}.
\bjtitle{Neuroimage}
\bvolume{9}(\bissue{2}),
\bfpage{179}--\blpage{194}
(\byear{1999})
\end{barticle}
\endbibitem

\bibitem[\protect\citeauthoryear{Fischl et~al.}{1999}]{fischl1999cortical}
\begin{barticle}
\bauthor{\bsnm{Fischl}, \binits{B.}},
\bauthor{\bsnm{Sereno}, \binits{M.}},
\bauthor{\bsnm{Dale}, \binits{A.M.}}:
\batitle{Cortical surface-based analysis: {II}: inflation, flattening, and a surface-based coordinate system}.
\bjtitle{Neuroimage}
\bvolume{9}(\bissue{2}),
\bfpage{195}--\blpage{207}
(\byear{1999})
\end{barticle}
\endbibitem

\bibitem[\protect\citeauthoryear{Tustison et~al.}{2014}]{tustison2014large}
\begin{barticle}
\bauthor{\bsnm{Tustison}, \binits{N.J.}},
\bauthor{\bsnm{Cook}, \binits{P.A.}},
\bauthor{\bsnm{Klein}, \binits{A.}},
\bauthor{\bsnm{Song}, \binits{G.}},
\bauthor{\bsnm{Das}, \binits{S.R.}},
\bauthor{\bsnm{Duda}, \binits{J.T.}},
\bauthor{\bsnm{Kandel}, \binits{B.M.}},
\bauthor{\bsnm{Strien}, \binits{N.}},
\bauthor{\bsnm{Stone}, \binits{J.R.}},
\bauthor{\bsnm{Gee}, \binits{J.C.}}, \betal:
\batitle{Large-scale evaluation of ants and freesurfer cortical thickness measurements}.
\bjtitle{Neuroimage}
\bvolume{99},
\bfpage{166}--\blpage{179}
(\byear{2014})
\end{barticle}
\endbibitem

\bibitem[\protect\citeauthoryear{Shattuck and Leahy}{2002}]{shattuck2002brainsuite}
\begin{barticle}
\bauthor{\bsnm{Shattuck}, \binits{D.W.}},
\bauthor{\bsnm{Leahy}, \binits{R.M.}}:
\batitle{Brainsuite: an automated cortical surface identification tool}.
\bjtitle{Medical image analysis}
\bvolume{6}(\bissue{2}),
\bfpage{129}--\blpage{142}
(\byear{2002})
\end{barticle}
\endbibitem

\bibitem[\protect\citeauthoryear{Lim et~al.}{2024}]{lim_low-field_2024}
\begin{barticle}
\bauthor{\bsnm{Lim}, \binits{T.R.}},
\bauthor{\bsnm{Suthiphosuwan}, \binits{S.}},
\bauthor{\bsnm{Micieli}, \binits{J.}},
\bauthor{\bsnm{Vosoughi}, \binits{R.}},
\bauthor{\bsnm{Schneider}, \binits{R.}},
\bauthor{\bsnm{Lin}, \binits{A.W.}},
\bauthor{\bsnm{Chen}, \binits{Y.A.}},
\bauthor{\bsnm{Muccilli}, \binits{A.}},
\bauthor{\bsnm{Marriott}, \binits{J.J.}},
\bauthor{\bsnm{Selchen}, \binits{D.}}, \betal:
\batitle{Low-field (64 mt) portable mri for rapid point-of-care diagnosis of dissemination in space in patients presenting with optic neuritis}.
\bjtitle{American Journal of Neuroradiology}
\bvolume{45}(\bissue{11}),
\bfpage{1819}--\blpage{1825}
(\byear{2024})
\end{barticle}
\endbibitem

\bibitem[\protect\citeauthoryear{Sheth et~al.}{2022}]{sheth_bedside_2022}
\begin{barticle}
\bauthor{\bsnm{Sheth}, \binits{K.N.}},
\bauthor{\bsnm{Yuen}, \binits{M.M.}},
\bauthor{\bsnm{Mazurek}, \binits{M.H.}},
\bauthor{\bsnm{Cahn}, \binits{B.A.}},
\bauthor{\bsnm{Prabhat}, \binits{A.M.}},
\bauthor{\bsnm{Salehi}, \binits{S.}},
\bauthor{\bsnm{Shah}, \binits{J.T.}},
\bauthor{\bsnm{By}, \binits{S.}},
\bauthor{\bsnm{Welch}, \binits{E.B.}},
\bauthor{\bsnm{Sofka}, \binits{M.}}, \betal:
\batitle{Bedside detection of intracranial midline shift using portable magnetic resonance imaging}.
\bjtitle{Scientific reports}
\bvolume{12}(\bissue{1}),
\bfpage{67}
(\byear{2022})
\end{barticle}
\endbibitem

\bibitem[\protect\citeauthoryear{Yuen et~al.}{2022}]{yuen2022portable}
\begin{barticle}
\bauthor{\bsnm{Yuen}, \binits{M.M.}},
\bauthor{\bsnm{Prabhat}, \binits{A.M.}},
\bauthor{\bsnm{Mazurek}, \binits{M.H.}},
\bauthor{\bsnm{Chavva}, \binits{I.R.}},
\bauthor{\bsnm{Crawford}, \binits{A.}},
\bauthor{\bsnm{Cahn}, \binits{B.A.}},
\bauthor{\bsnm{Beekman}, \binits{R.}},
\bauthor{\bsnm{Kim}, \binits{J.A.}},
\bauthor{\bsnm{Gobeske}, \binits{K.T.}},
\bauthor{\bsnm{Petersen}, \binits{N.H.}}, \betal:
\batitle{Portable, low-field magnetic resonance imaging enables highly accessible and dynamic bedside evaluation of ischemic stroke}.
\bjtitle{Science advances}
\bvolume{8}(\bissue{16}),
\bfpage{3952}
(\byear{2022})
\end{barticle}
\endbibitem

\bibitem[\protect\citeauthoryear{Arnold et~al.}{2023}]{arnold_low-field_2023}
\begin{barticle}
\bauthor{\bsnm{Arnold}, \binits{T.C.}},
\bauthor{\bsnm{Freeman}, \binits{C.W.}},
\bauthor{\bsnm{Litt}, \binits{B.}},
\bauthor{\bsnm{Stein}, \binits{J.M.}}:
\batitle{Low-field mri: clinical promise and challenges}.
\bjtitle{Journal of Magnetic Resonance Imaging}
\bvolume{57}(\bissue{1}),
\bfpage{25}--\blpage{44}
(\byear{2023})
\end{barticle}
\endbibitem

\bibitem[\protect\citeauthoryear{Laso et~al.}{2024}]{laso2024quantifying}
\begin{bchapter}
\bauthor{\bsnm{Laso}, \binits{P.}},
\bauthor{\bsnm{Cerri}, \binits{S.}},
\bauthor{\bsnm{Sorby-Adams}, \binits{A.}},
\bauthor{\bsnm{Guo}, \binits{J.}},
\bauthor{\bsnm{Mateen}, \binits{F.}},
\bauthor{\bsnm{Goebl}, \binits{P.}},
\bauthor{\bsnm{Wu}, \binits{J.}},
\bauthor{\bsnm{Liu}, \binits{P.}},
\bauthor{\bsnm{Li}, \binits{H.B.}},
\bauthor{\bsnm{Young}, \binits{S.I.}}, \betal:
\bctitle{Quantifying white matter hyperintensity and brain volumes in heterogeneous clinical and low-field portable mri}.
In: \bbtitle{2024 IEEE International Symposium on Biomedical Imaging (ISBI)},
pp. \bfpage{1}--\blpage{5}
(\byear{2024}).
\bcomment{IEEE}
\end{bchapter}
\endbibitem

\bibitem[\protect\citeauthoryear{Shay et~al.}{2024}]{shay2024portable}
\begin{bchapter}
\bauthor{\bsnm{Shay}, \binits{J.}},
\bauthor{\bsnm{Nimjee}, \binits{S.}},
\bauthor{\bsnm{Forrest}, \binits{C.}},
\bauthor{\bsnm{Shujaat}, \binits{M.T.}},
\bauthor{\bsnm{Morales}, \binits{J.}},
\bauthor{\bsnm{Lee}, \binits{V.}}:
\bctitle{Portable bedside low-field magnetic resonance imaging for evaluation of ischemic strokes in patients admitted to inpatient floor units (p3-5.012)}.
In: \bbtitle{Neurology},
vol. \bseriesno{102},
p. \bfpage{3445}
(\byear{2024})
\end{bchapter}
\endbibitem

\bibitem[\protect\citeauthoryear{Zabinska et~al.}{2024}]{zabinska2024low}
\begin{bchapter}
\bauthor{\bsnm{Zabinska}, \binits{J.}},
\bauthor{\bsnm{Yadlapalli}, \binits{V.}},
\bauthor{\bsnm{Mazurek}, \binits{M.}},
\bauthor{\bsnm{Parasuram}, \binits{N.}},
\bauthor{\bsnm{Lalwani}, \binits{D.}},
\bauthor{\bsnm{Peasley}, \binits{E.}},
\bauthor{\bsnm{Gilmore}, \binits{E.}},
\bauthor{\bsnm{Kim}, \binits{J.}},
\bauthor{\bsnm{Omay}, \binits{S.B.}},
\bauthor{\bsnm{Sze}, \binits{G.}}, \betal:
\bctitle{Low-field, portable magnetic resonance imaging to assess hematoma volume in traumatic brain injury patients (p10-5.008)}.
In: \bbtitle{Neurology},
vol. \bseriesno{102},
p. \bfpage{6974}
(\byear{2024})
\end{bchapter}
\endbibitem

\bibitem[\protect\citeauthoryear{Sorby-Adams et~al.}{2024}]{sorby2024portable}
\begin{barticle}
\bauthor{\bsnm{Sorby-Adams}, \binits{A.J.}},
\bauthor{\bsnm{Guo}, \binits{J.}},
\bauthor{\bsnm{Laso}, \binits{P.}},
\bauthor{\bsnm{Kirsch}, \binits{J.E.}},
\bauthor{\bsnm{Zabinska}, \binits{J.}},
\bauthor{\bsnm{Garcia~Guarniz}, \binits{A.-L.}},
\bauthor{\bsnm{Schaefer}, \binits{P.W.}},
\bauthor{\bsnm{Payabvash}, \binits{S.}},
\bauthor{\bsnm{Havenon}, \binits{A.}},
\bauthor{\bsnm{Rosen}, \binits{M.S.}}, \betal:
\batitle{Portable, low-field magnetic resonance imaging for evaluation of alzheimer’s disease}.
\bjtitle{Nature Communications}
\bvolume{15}(\bissue{1}),
\bfpage{1}--\blpage{12}
(\byear{2024})
\end{barticle}
\endbibitem

\bibitem[\protect\citeauthoryear{Henschel et~al.}{2022}]{henschel2022fastsurfervinn}
\begin{barticle}
\bauthor{\bsnm{Henschel}, \binits{L.}},
\bauthor{\bsnm{K{\"u}gler}, \binits{D.}},
\bauthor{\bsnm{Reuter}, \binits{M.}}:
\batitle{Fastsurfervinn: Building resolution-independence into deep learning segmentation methods—a solution for highres brain mri}.
\bjtitle{NeuroImage}
\bvolume{251},
\bfpage{118933}
(\byear{2022})
\end{barticle}
\endbibitem

\bibitem[\protect\citeauthoryear{Bongratz et~al.}{2022}]{bongratz2022vox2cortex}
\begin{bchapter}
\bauthor{\bsnm{Bongratz}, \binits{F.}},
\bauthor{\bsnm{Rickmann}, \binits{A.-M.}},
\bauthor{\bsnm{P{\"o}lsterl}, \binits{S.}},
\bauthor{\bsnm{Wachinger}, \binits{C.}}:
\bctitle{{Vox2Cortex}: Fast explicit reconstruction of cortical surfaces from {3D} {MRI} scans with geometric deep neural networks}.
In: \bbtitle{CVPR},
pp. \bfpage{20773}--\blpage{20783}
(\byear{2022})
\end{bchapter}
\endbibitem

\bibitem[\protect\citeauthoryear{Cruz et~al.}{2021}]{cruz2021deepcsr}
\begin{bchapter}
\bauthor{\bsnm{Cruz}, \binits{R.S.}},
\bauthor{\bsnm{Lebrat}, \binits{L.}},
\bauthor{\bsnm{Bourgeat}, \binits{P.}},
\bauthor{\bsnm{Fookes}, \binits{C.}},
\bauthor{\bsnm{Fripp}, \binits{J.}},
\bauthor{\bsnm{Salvado}, \binits{O.}}:
\bctitle{Deep{CSR}: A {3D} deep learning approach for cortical surface reconstruction}.
In: \bbtitle{WACV},
pp. \bfpage{806}--\blpage{815}
(\byear{2021})
\end{bchapter}
\endbibitem

\bibitem[\protect\citeauthoryear{Gopinath et~al.}{2023}]{gopinath2023cortical}
\begin{bchapter}
\bauthor{\bsnm{Gopinath}, \binits{K.}},
\bauthor{\bsnm{Greve}, \binits{D.N.}},
\bauthor{\bsnm{Das}, \binits{S.}},
\bauthor{\bsnm{Arnold}, \binits{S.}},
\bauthor{\bsnm{Magdamo}, \binits{C.}},
\bauthor{\bsnm{Iglesias}, \binits{J.E.}}:
\bctitle{Cortical analysis of heterogeneous clinical brain mri scans for large-scale neuroimaging studies}.
In: \bbtitle{International Conference on Medical Image Computing and Computer-Assisted Intervention},
pp. \bfpage{35}--\blpage{45}
(\byear{2023}).
\bcomment{Springer}
\end{bchapter}
\endbibitem

\bibitem[\protect\citeauthoryear{Ma et~al.}{2022}]{ma2022cortexode}
\begin{barticle}
\bauthor{\bsnm{Ma}, \binits{Q.}},
\bauthor{\bsnm{Li}, \binits{L.}},
\bauthor{\bsnm{Robinson}, \binits{E.C.}},
\bauthor{\bsnm{Kainz}, \binits{B.}},
\bauthor{\bsnm{Rueckert}, \binits{D.}},
\bauthor{\bsnm{Alansary}, \binits{A.}}:
\batitle{{CortexODE}: Learning cortical surface reconstruction by neural {ODEs}}.
\bjtitle{IEEE Transactions on Medical Imaging}
\bvolume{42}(\bissue{2}),
\bfpage{430}--\blpage{443}
(\byear{2022})
\end{barticle}
\endbibitem

\bibitem[\protect\citeauthoryear{Hoopes et~al.}{2022}]{hoopes2022topofit}
\begin{barticle}
\bauthor{\bsnm{Hoopes}, \binits{A.}},
\bauthor{\bsnm{Iglesias}, \binits{J.E.}},
\bauthor{\bsnm{Fischl}, \binits{B.}},
\bauthor{\bsnm{Greve}, \binits{D.}},
\bauthor{\bsnm{Dalca}, \binits{A.V.}}:
\batitle{{TopoFit}: rapid reconstruction of topologically-correct cortical surfaces}.
\bjtitle{Proceedings of machine learning research}
\bvolume{172},
\bfpage{508}
(\byear{2022})
\end{barticle}
\endbibitem

\bibitem[\protect\citeauthoryear{Iglesias et~al.}{2023}]{iglesias2023synthsr}
\begin{barticle}
\bauthor{\bsnm{Iglesias}, \binits{J.E.}},
\bauthor{\bsnm{Billot}, \binits{B.}},
\bauthor{\bsnm{Balbastre}, \binits{Y.}},
\bauthor{\bsnm{Magdamo}, \binits{C.}},
\bauthor{\bsnm{Arnold}, \binits{S.E.}},
\bauthor{\bsnm{Das}, \binits{S.}},
\bauthor{\bsnm{Edlow}, \binits{B.L.}},
\bauthor{\bsnm{Alexander}, \binits{D.C.}},
\bauthor{\bsnm{Golland}, \binits{P.}},
\bauthor{\bsnm{Fischl}, \binits{B.}}:
\batitle{{SynthSR}: A public {AI} tool to turn heterogeneous clinical brain scans into high-resolution {T1}-weighted images for {3D} morphometry}.
\bjtitle{Science advances}
\bvolume{9}(\bissue{5}),
\bfpage{3607}
(\byear{2023})
\end{barticle}
\endbibitem

\bibitem[\protect\citeauthoryear{Iglesias et~al.}{2022}]{iglesias2022quantitative}
\begin{barticle}
\bauthor{\bsnm{Iglesias}, \binits{J.E.}},
\bauthor{\bsnm{Schleicher}, \binits{R.}},
\bauthor{\bsnm{Laguna}, \binits{S.}},
\bauthor{\bsnm{Billot}, \binits{B.}},
\bauthor{\bsnm{Schaefer}, \binits{P.}},
\bauthor{\bsnm{McKaig}, \binits{B.}},
\bauthor{\bsnm{Goldstein}, \binits{J.N.}},
\bauthor{\bsnm{Sheth}, \binits{K.N.}},
\bauthor{\bsnm{Rosen}, \binits{M.S.}},
\bauthor{\bsnm{Kimberly}, \binits{W.T.}}:
\batitle{Quantitative brain morphometry of portable low-field-strength {MRI} using super-resolution machine learning}.
\bjtitle{Radiology}
\bvolume{306}(\bissue{3}),
\bfpage{220522}
(\byear{2022})
\end{barticle}
\endbibitem

\bibitem[\protect\citeauthoryear{Desikan et~al.}{2006}]{desikan2006automated}
\begin{barticle}
\bauthor{\bsnm{Desikan}, \binits{R.S.}},
\bauthor{\bsnm{S{\'e}gonne}, \binits{F.}},
\bauthor{\bsnm{Fischl}, \binits{B.}},
\bauthor{\bsnm{Quinn}, \binits{B.T.}},
\bauthor{\bsnm{Dickerson}, \binits{B.C.}},
\bauthor{\bsnm{Blacker}, \binits{D.}},
\bauthor{\bsnm{Buckner}, \binits{R.L.}},
\bauthor{\bsnm{Dale}, \binits{A.M.}}, \betal:
\batitle{An automated labeling system for subdividing the human cerebral cortex on mri scans into gyral based regions of interest}.
\bjtitle{NeuroImage}
\bvolume{31}(\bissue{3}),
\bfpage{968}--\blpage{980}
(\byear{2006})
\end{barticle}
\endbibitem

\bibitem[\protect\citeauthoryear{Gopinath et~al.}{2024}]{gopinath2024recon}
\begin{botherref}
\oauthor{\bsnm{Gopinath}, \binits{K.}},
\oauthor{\bsnm{Greve}, \binits{D.N.}},
\oauthor{\bsnm{Magdamo}, \binits{C.}},
\oauthor{\bsnm{Arnold}, \binits{S.}},
\oauthor{\bsnm{Das}, \binits{S.}},
\oauthor{\bsnm{Puonti}, \binits{O.}},
\oauthor{\bsnm{Iglesias}, \binits{J.E.}}:
Recon-all-clinical: Cortical surface reconstruction and analysis of heterogeneous clinical brain mri.
arXiv preprint arXiv:2409.03889
(2024)
\end{botherref}
\endbibitem

\bibitem[\protect\citeauthoryear{Glasser et~al.}{2013}]{glasser2013minimal}
\begin{barticle}
\bauthor{\bsnm{Glasser}, \binits{M.}},
\bauthor{\bsnm{Sotiropoulos}, \binits{S.}},
\bauthor{\bsnm{Wilson}, \binits{J.A.}},
\bauthor{\bsnm{Coalson}, \binits{T.S.}}:
\batitle{The minimal preprocessing pipelines for the {Human Connectome Project}}.
\bjtitle{Neuroimage}
\bvolume{80},
\bfpage{105}--\blpage{24}
(\byear{2013})
\end{barticle}
\endbibitem

\bibitem[\protect\citeauthoryear{Jack~Jr et~al.}{2008}]{jack2008alzheimer}
\begin{barticle}
\bauthor{\bsnm{Jack~Jr}, \binits{C.R.}},
\bauthor{\bsnm{Bernstein}, \binits{M.A.}},
\bauthor{\bsnm{Fox}, \binits{N.C.}},
\bauthor{\bsnm{Thompson}, \binits{P.}},
\bauthor{\bsnm{Alexander}, \binits{G.}},
\bauthor{\bsnm{Harvey}, \binits{D.}},
\bauthor{\bsnm{Borowski}, \binits{B.}},
\bauthor{\bsnm{Britson}, \binits{P.J.}},
\bauthor{\bsnm{L.~Whitwell}, \binits{J.}},
\bauthor{\bsnm{Ward}, \binits{C.}}, \betal:
\batitle{The {Alzheimer's} disease neuroimaging initiative ({ADNI}): {MRI} methods}.
\bjtitle{Journal of Magnetic Resonance Imaging: An Official Journal of the International Society for Magnetic Resonance in Medicine}
\bvolume{27}(\bissue{4}),
\bfpage{685}--\blpage{691}
(\byear{2008})
\end{barticle}
\endbibitem

\bibitem[\protect\citeauthoryear{Puonti et~al.}{2016}]{PUONTI2016235}
\begin{barticle}
\bauthor{\bsnm{Puonti}, \binits{O.}},
\bauthor{\bsnm{Iglesias}, \binits{J.E.}},
\bauthor{\bsnm{{Van Leemput}}, \binits{K.}}:
\batitle{Fast and sequence-adaptive whole-brain segmentation using parametric bayesian modeling}.
\bjtitle{NeuroImage}
\bvolume{143},
\bfpage{235}--\blpage{249}
(\byear{2016})
\end{barticle}
\endbibitem

\bibitem[\protect\citeauthoryear{Gopinath et~al.}{2024}]{gopinath2024synthetic}
\begin{barticle}
\bauthor{\bsnm{Gopinath}, \binits{K.}},
\bauthor{\bsnm{Hoopes}, \binits{A.}},
\bauthor{\bsnm{Alexander}, \binits{D.C.}},
\bauthor{\bsnm{Arnold}, \binits{S.E.}},
\bauthor{\bsnm{Balbastre}, \binits{Y.}},
\bauthor{\bsnm{Billot}, \binits{B.}},
\bauthor{\bsnm{Casamitjana}, \binits{A.}},
\bauthor{\bsnm{Cheng}, \binits{Y.}},
\bauthor{\bsnm{Chua}, \binits{R.Y.Z.}}, \betal:
\batitle{Synthetic data in generalizable, learning-based neuroimaging}.
\bjtitle{Imaging Neuroscience}
\bvolume{2},
\bfpage{1}--\blpage{22}
(\byear{2024})
\end{barticle}
\endbibitem

\bibitem[\protect\citeauthoryear{Billot et~al.}{2023}]{billot2023synthseg}
\begin{barticle}
\bauthor{\bsnm{Billot}, \binits{B.}},
\bauthor{\bsnm{Greve}, \binits{D.N.}},
\bauthor{\bsnm{Puonti}, \binits{O.}},
\bauthor{\bsnm{Thielscher}, \binits{A.}},
\bauthor{\bsnm{Van~Leemput}, \binits{K.}},
\bauthor{\bsnm{Fischl}, \binits{B.}},
\bauthor{\bsnm{Dalca}, \binits{A.V.}},
\bauthor{\bsnm{Iglesias}, \binits{J.E.}}, \betal:
\batitle{Synthseg: Segmentation of brain mri scans of any contrast and resolution without retraining}.
\bjtitle{Medical image analysis}
\bvolume{86},
\bfpage{102789}
(\byear{2023})
\end{barticle}
\endbibitem

\bibitem[\protect\citeauthoryear{{\c{C}}i{\c{c}}ek et~al.}{2016}]{cciccek20163d}
\begin{bchapter}
\bauthor{\bsnm{{\c{C}}i{\c{c}}ek}, \binits{{\"O}.}},
\bauthor{\bsnm{Abdulkadir}, \binits{A.}},
\bauthor{\bsnm{Lienkamp}, \binits{S.S.}},
\bauthor{\bsnm{Brox}, \binits{T.}},
\bauthor{\bsnm{Ronneberger}, \binits{O.}}:
\bctitle{3d u-net: learning dense volumetric segmentation from sparse annotation}.
In: \bbtitle{Medical Image Computing and Computer-Assisted Intervention--MICCAI 2016: 19th International Conference, Athens, Greece, October 17-21, 2016, Proceedings, Part II 19},
pp. \bfpage{424}--\blpage{432}
(\byear{2016})
\end{bchapter}
\endbibitem

\bibitem[\protect\citeauthoryear{Kingma and Ba}{2014}]{kingma2014adam}
\begin{botherref}
\oauthor{\bsnm{Kingma}, \binits{D.P.}},
\oauthor{\bsnm{Ba}, \binits{J.}}:
Adam: A method for stochastic optimization.
arXiv preprint arXiv:1412.6980
(2014)
\end{botherref}
\endbibitem

\bibitem[\protect\citeauthoryear{Lorensen and Cline}{1998}]{lorensen1998marching}
\begin{bchapter}
\bauthor{\bsnm{Lorensen}, \binits{W.E.}},
\bauthor{\bsnm{Cline}, \binits{H.E.}}:
\bctitle{Marching cubes: A high resolution 3d surface construction algorithm}.
In: \bbtitle{Seminal Graphics: Pioneering Efforts that Shaped the Field},
pp. \bfpage{347}--\blpage{353}
(\byear{1998})
\end{bchapter}
\endbibitem

\bibitem[\protect\citeauthoryear{Fischl}{2012}]{fischl2012freesurfer}
\begin{barticle}
\bauthor{\bsnm{Fischl}, \binits{B.}}:
\batitle{Freesurfer}.
\bjtitle{NeuroImage}
\bvolume{62}(\bissue{2}),
\bfpage{774}--\blpage{781}
(\byear{2012})
\end{barticle}
\endbibitem

\bibitem[\protect\citeauthoryear{Fischl et~al.}{2004}]{fischl2004automatically}
\begin{barticle}
\bauthor{\bsnm{Fischl}, \binits{B.}},
\bauthor{\bsnm{Kouwe}, \binits{A.V.D.}},
\bauthor{\bsnm{Destrieux}, \binits{C.}},
\bauthor{\bsnm{Halgren}, \binits{E.}},
\bauthor{\bsnm{S{\'e}gonne}, \binits{F.}},
\bauthor{\bsnm{Salat}, \binits{D.H.}},
\bauthor{\bsnm{Busa}, \binits{E.}},
\bauthor{\bsnm{Seidman}, \binits{L.J.}},
\bauthor{\bsnm{Goldstein}, \binits{J.}},
\bauthor{\bsnm{Kennedy}, \binits{D.}}, \betal:
\batitle{Automatically parcellating the human cerebral cortex}.
\bjtitle{Cerebral Cortex}
\bvolume{14}(\bissue{1}),
\bfpage{11}--\blpage{22}
(\byear{2004})
\end{barticle}
\endbibitem

\end{thebibliography}

\newpage
\appendix
    \section{LF-MRI Sequence Parameters and Acquisition Times}
        \setcounter{table}{0}
        \renewcommand{\thetable}{S\arabic{table}} 
        \begin{table}[h]
        \small
        \setlength{\tabcolsep}{5pt}
        \centering
        \caption{LF-MRI (0.064 T) sequence parameters and acquisition times.}
        \label{tab:lfmri-params}
        \begin{tabular}{lcccccc}
        \toprule
        \textbf{Sequence} & \textbf{Time} & \textbf{TR} & \textbf{TE} & \textbf{TI} & \textbf{Pixel} & \textbf{Slice} \\
        & \textbf{(min:sec)} & \textbf{(ms)} & \textbf{(ms)} & \textbf{(ms)} & \textbf{Spacing (mm)} & \textbf{Thickness (mm)} \\
        \midrule
        \multicolumn{7}{l}{\textbf{LF T1w}} \\
        Axial In-Plane & 3:41 & 880 & 7.1 & 354 & 1.6 & 5 \\
        2.0 mm Isotropic & 9:48 & 880 & 6.19 & 354 & 2 & 2 \\
        3.0 mm Isotropic & 4:27 & 880 & 5.02 & 354 & 3 & 3 \\
        4.0 mm Isotropic & 2:35 & 880 & 4.33 & 354 & 3 & 3 \\
        \multicolumn{7}{l}{\textbf{LF T2w}} \\
        Axial In-Plane & 2:37 & 2000 & 180.8 & N/A & 1.6 & 5 \\
        2.0 mm Isotropic & 6:47 & 2000 & 210 & N/A & 2 & 2 \\
        3.0 mm Isotropic & 3:09 & 2000 & 169.2 & N/A & 3 & 3 \\
        4.0 mm Isotropic & 1:53 & 2000 & 145.2 & N/A & 4 & 4 \\
        \multicolumn{7}{l}{\textbf{Postmortem LF T2w}} \\
        Fresh and Cadeveric & 10:02 & 2000 & 183.2 & N/A & 2.30 & 2.31 \\
        \bottomrule
        \end{tabular}
        \end{table}
        
    \newpage
    \section{Cortical Parcellation Accuracy Across Resolutions}
        \setcounter{figure}{0}
        \renewcommand{\thefigure}{F\arabic{figure}} 
        
        \begin{figure*}[h]
            \centering
            {\bfseries Axial (1.6~mm $\times$ 1.6~mm $\times$ 5~mm)}\\[0.5ex]
            \includegraphics[width=\textwidth]{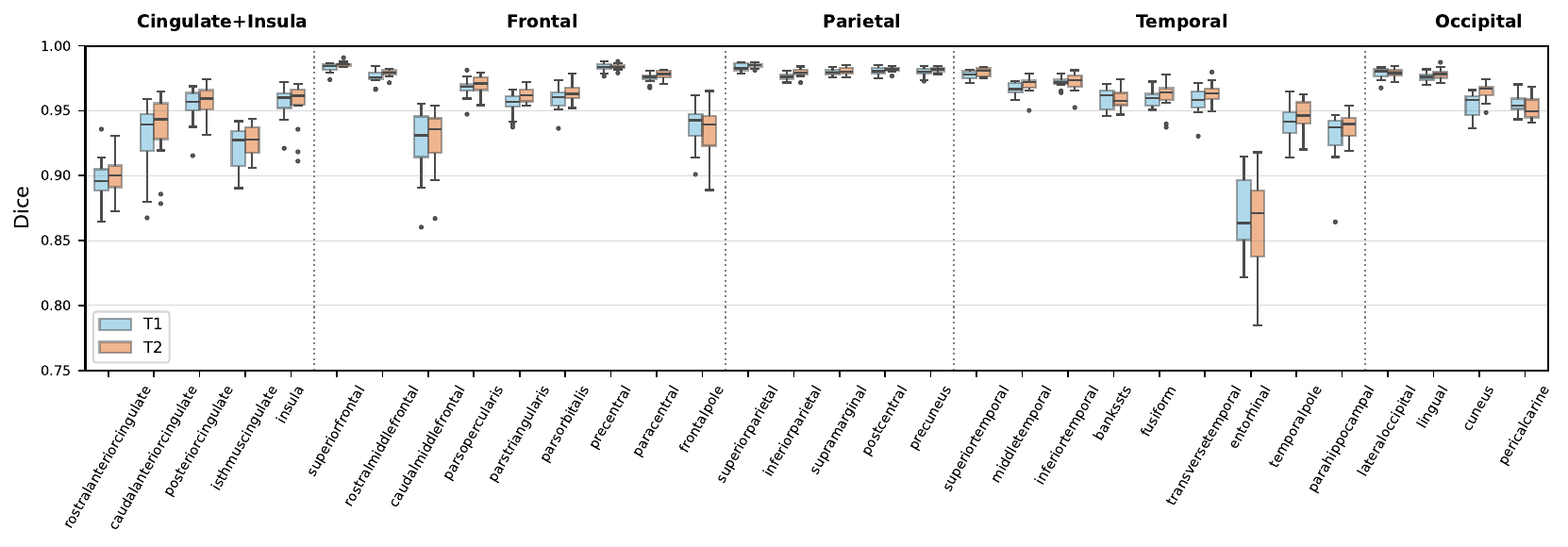} \\[2ex]
        
            {\bfseries 2~mm Isotropic}\\[0.5ex]
            \includegraphics[width=\textwidth]{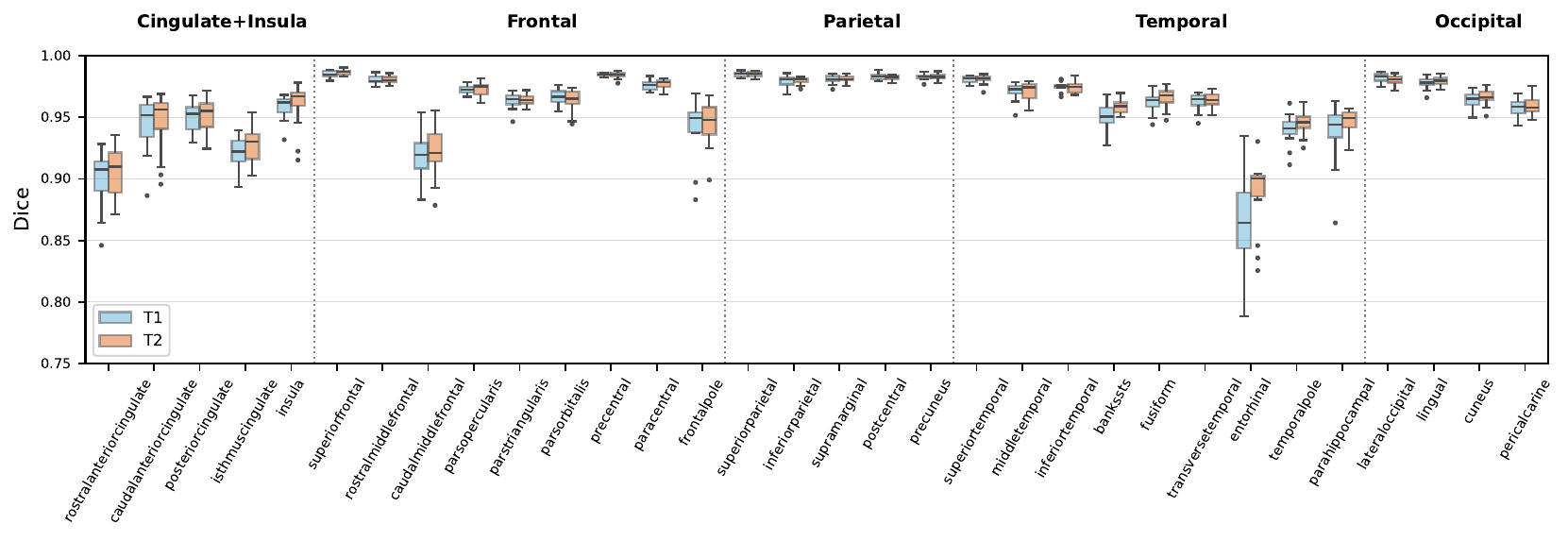} \\[2ex]
        
            \caption{Cortical parcellation accuracy for axial and 2~mm isotropic LF-MRI scans. Dice similarity coefficients (DSC) between LF-MRI-derived parcellations and HF-MRI reference are grouped by lobe and shown for T1 (blue) and T2 (orange).}
            \label{fig:cor_parc_app1}
        \end{figure*}
     \newpage   
        \begin{figure*}[h]
            \centering
            {\bfseries 3~mm Isotropic}\\[0.5ex]
            \includegraphics[width=\textwidth]{figures/combined_boxplot_parcel_lobewise_3mm.pdf} \\[2ex]
        
            {\bfseries 4~mm Isotropic}\\[0.5ex]
            \includegraphics[width=\textwidth]{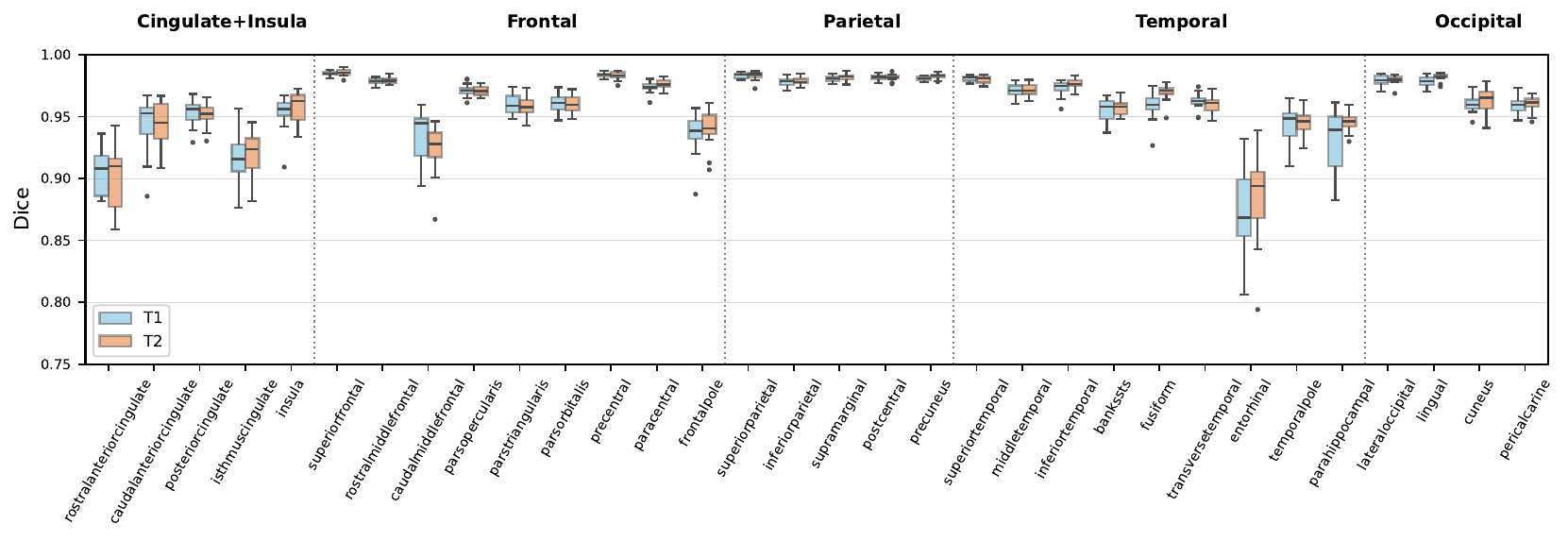} \\[2ex]
        
            \caption{Cortical parcellation accuracy for 3~mm and 4~mm isotropic LF-MRI scans. Dice similarity coefficients (DSC) between LF-MRI-derived parcellations and HF-MRI reference are grouped by lobe and shown for T1 (blue) and T2 (orange).}
            \label{fig:cor_parc_app2}
        \end{figure*}

    \newpage
    \section{Cortical Morphometry Across Resolutions}
        \setcounter{figure}{2}
        \renewcommand{\thefigure}{F\arabic{figure}} 
        \begin{figure*}[h]
            \centering
            {\bfseries Axial (1.6~mm $\times$ 1.6~mm $\times$ 5~mm)}
            \includegraphics[width=0.9\textwidth]{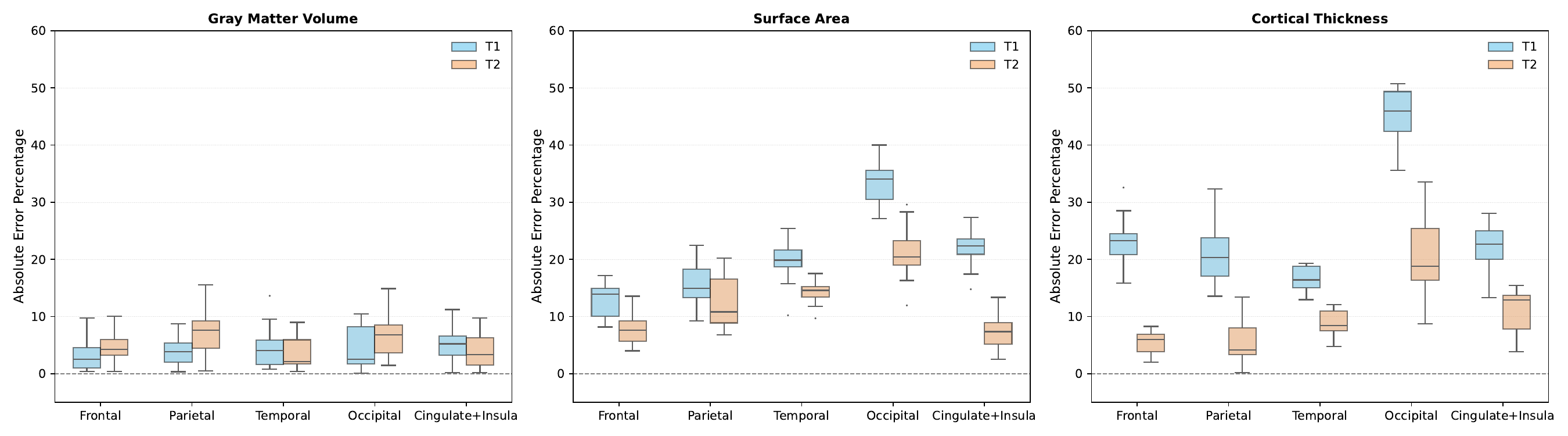} 
        
            {\bfseries 2~mm Isotropic}
            \includegraphics[width=0.9\textwidth]{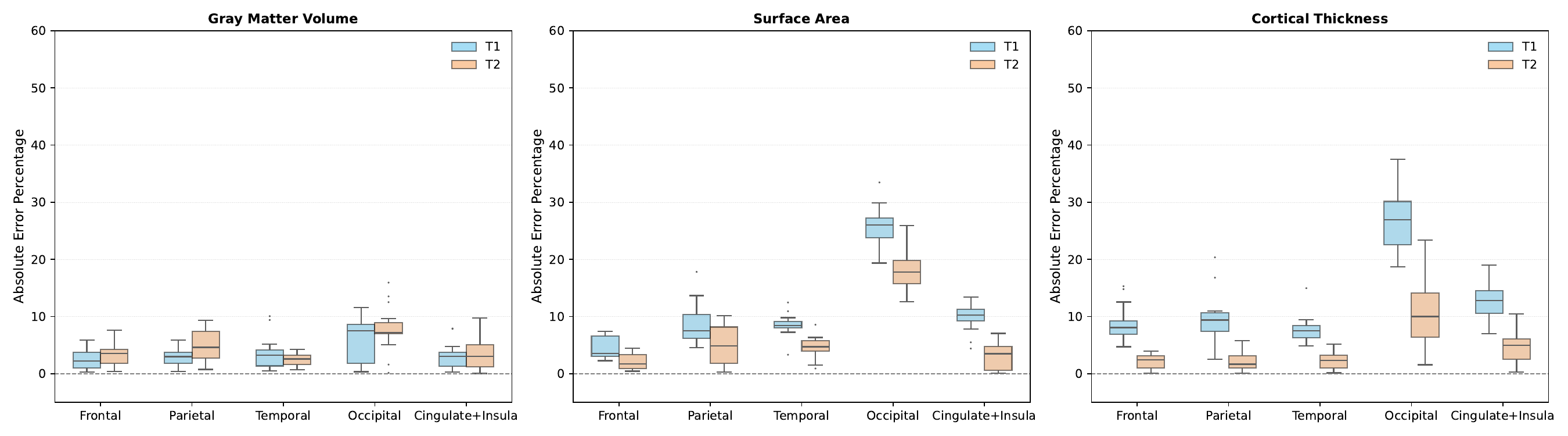} 
        
            {\bfseries 3~mm Isotropic}
            \includegraphics[width=0.9\textwidth]{figures/lobar_differences_gt_pred_pretty_percentatge_3mm.pdf} 
        
            {\bfseries 4~mm Isotropic}
            \includegraphics[width=0.9\textwidth]{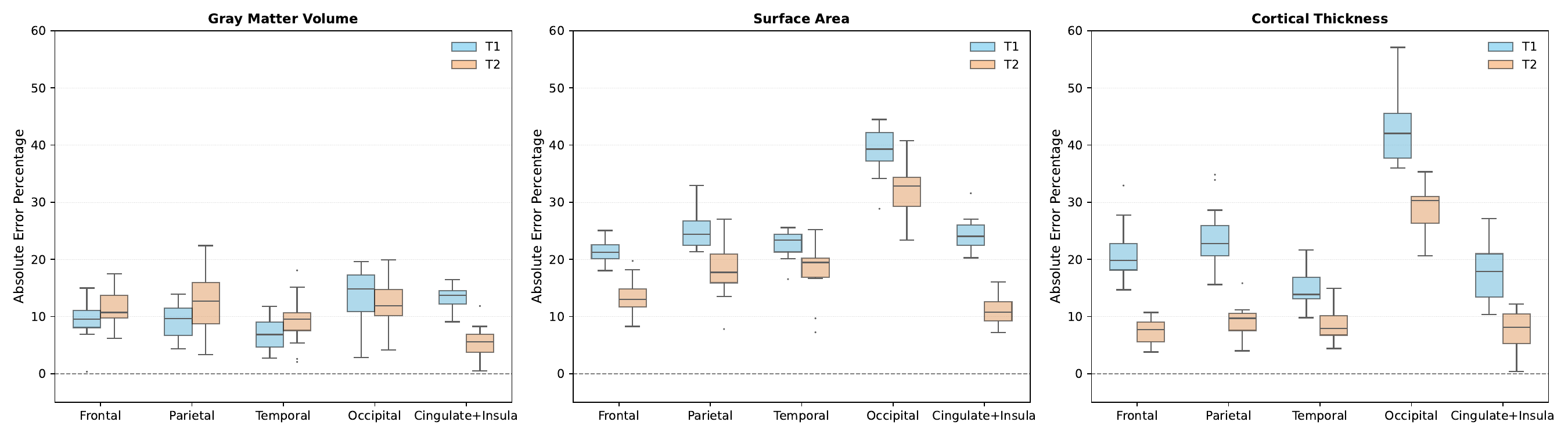} 
        
            \caption{Absolute percentage differences in surface area, gray matter volume, and cortical thickness across anatomical regions, relative to HF-MRI reference. Results are shown for T1 (blue) and T2 (orange) scans.}

            \label{fig:cort_morpho_allres}
        \end{figure*}

    \newpage
    \section{Qualitative Validation on Postmortem LF-MRI}
        
        \begin{figure*}[ht]
            \centering
            {\bfseries Postmortem cortical reconstruction from LF-MRI scans.}\\
            \includegraphics[width=\textwidth]{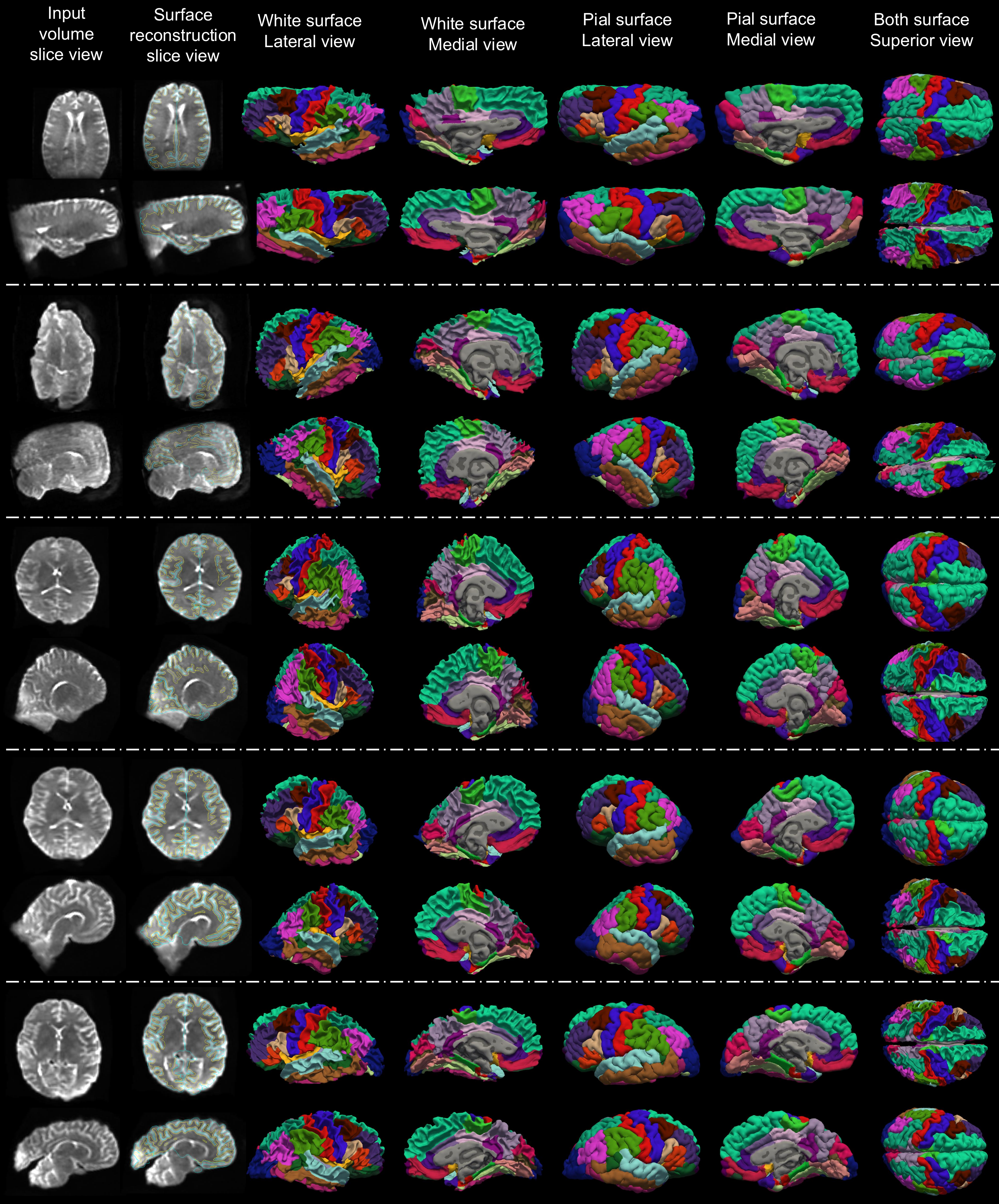}
        \end{figure*}
        \begin{figure*}[ht]
            \centering
            \includegraphics[width=\textwidth]{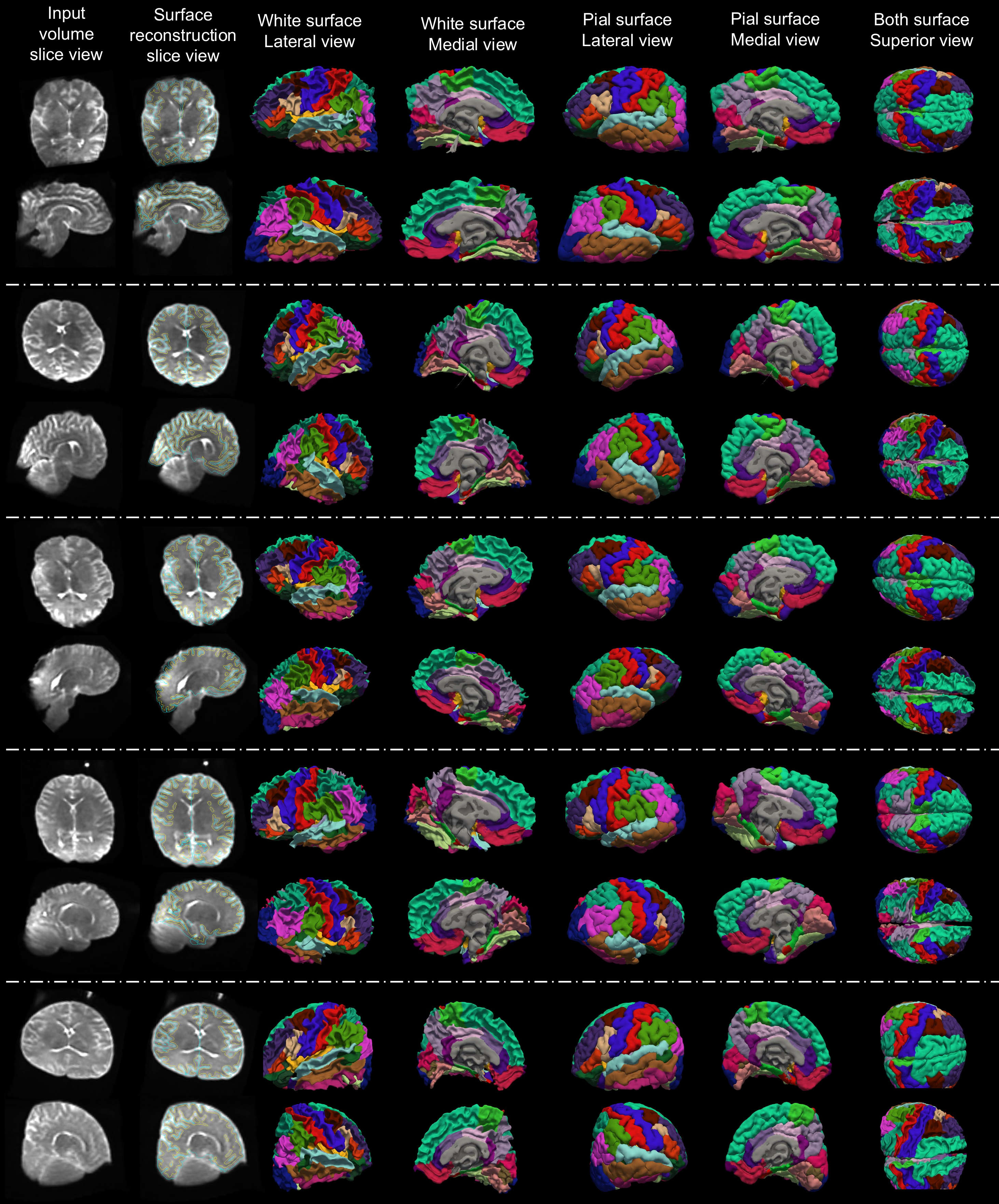}
        \end{figure*}
        \begin{figure*}[ht]
            \centering
            \includegraphics[width=\textwidth]{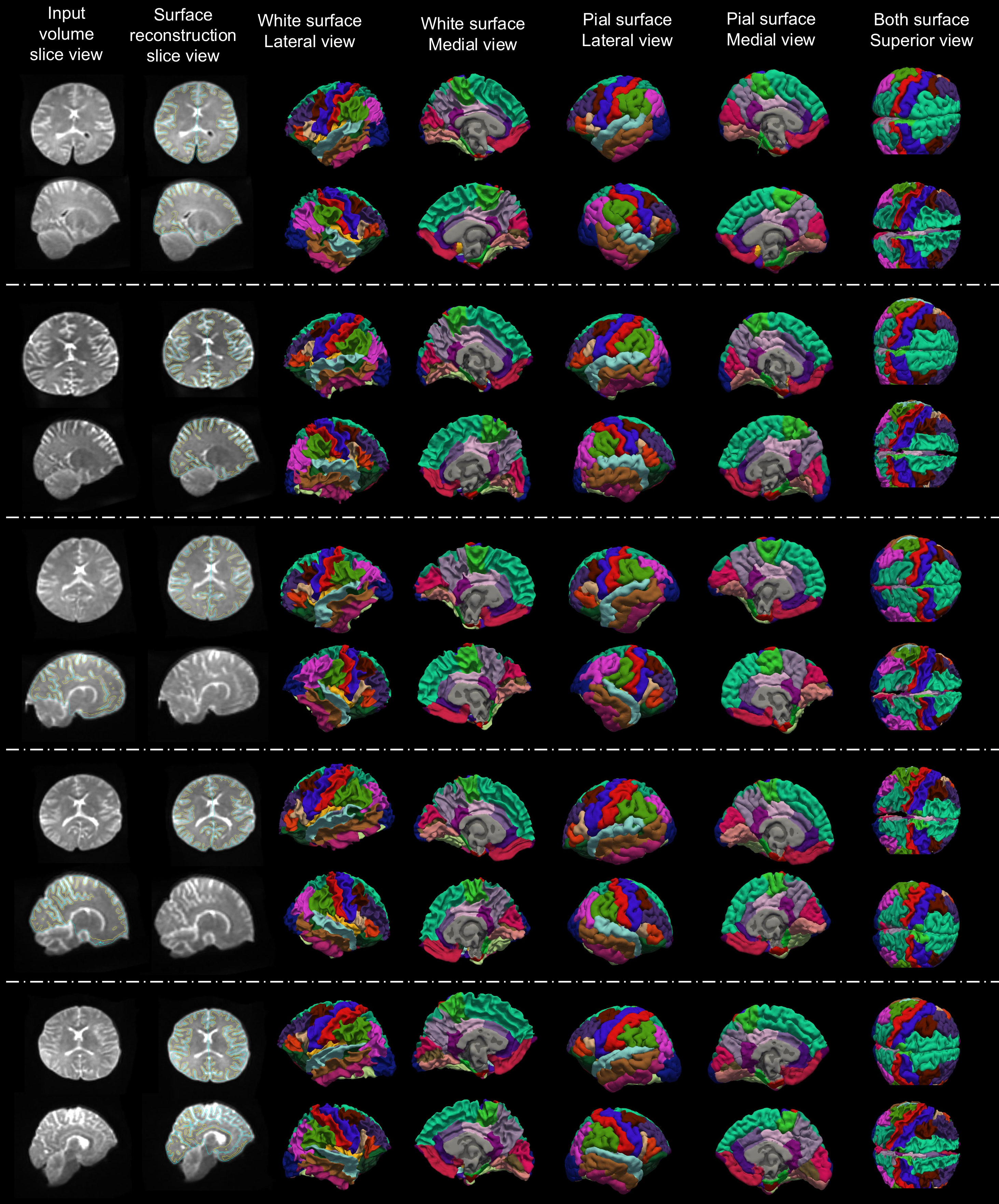}
        \end{figure*}
        \begin{figure*}[ht]
            \centering
            \includegraphics[width=\textwidth]{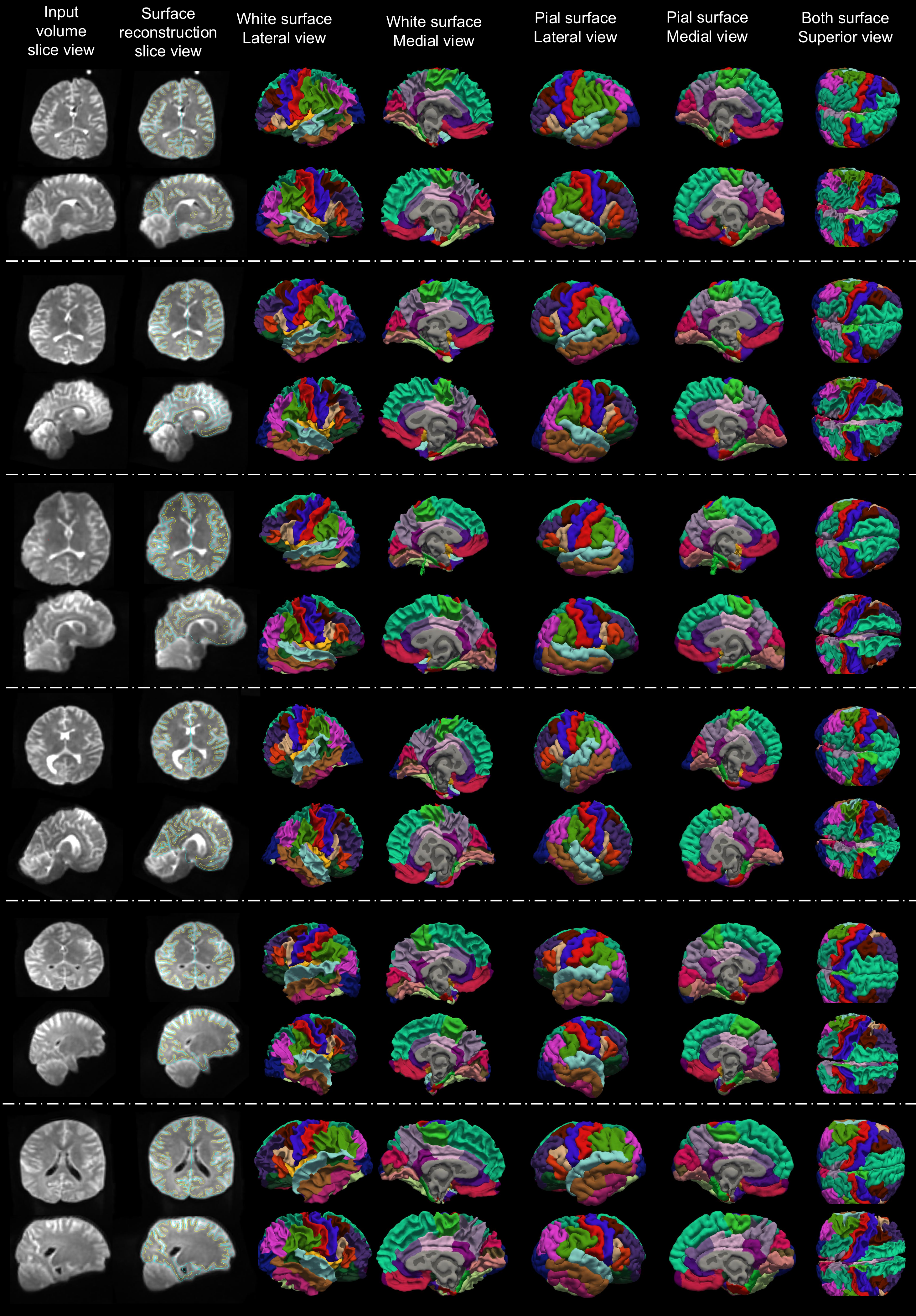}
        \end{figure*}
        
        \begin{figure*}[ht]
            \centering
            \includegraphics[width=\textwidth]{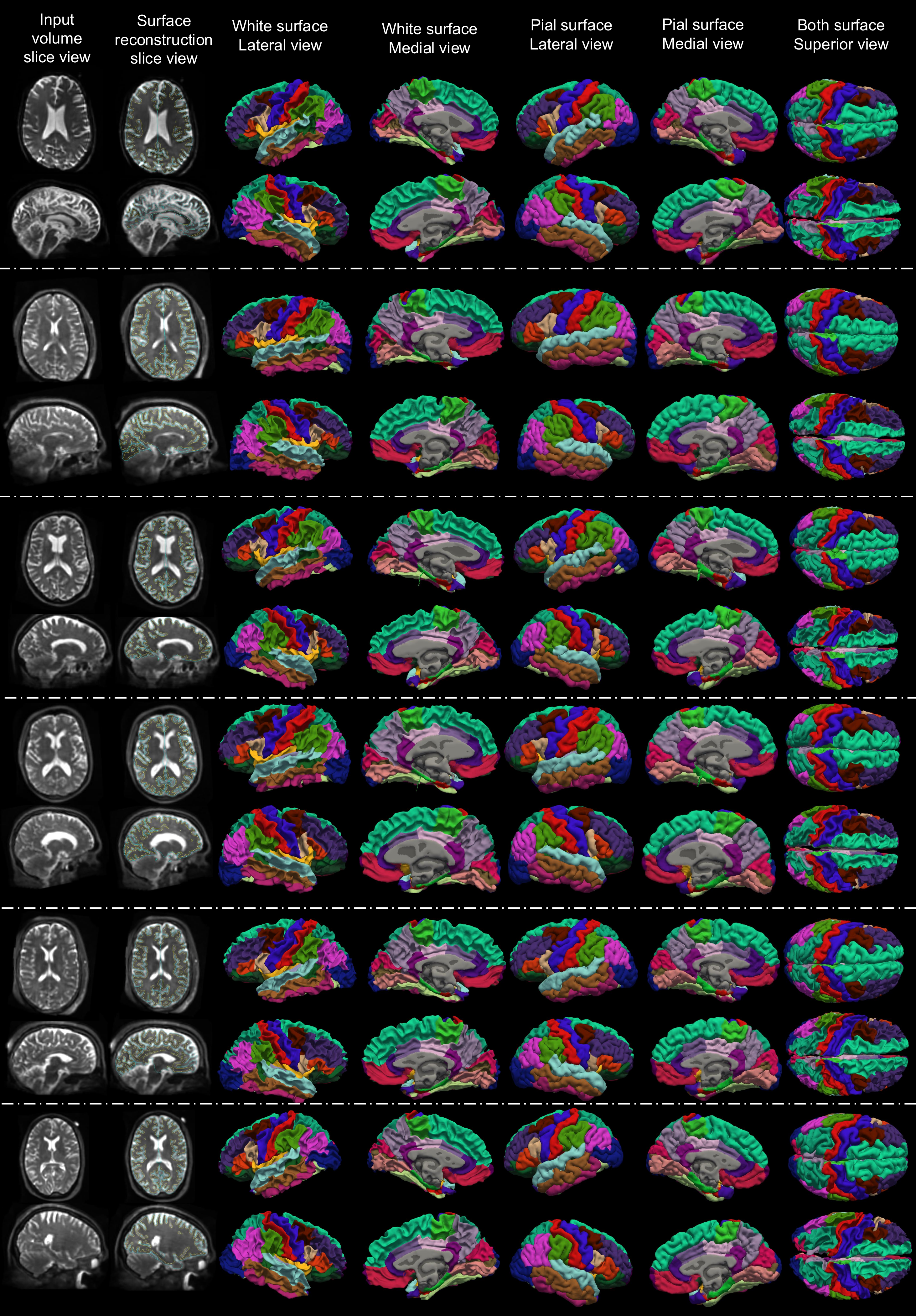}
        \end{figure*}
        \begin{figure*}[ht]
            \centering
        
            \includegraphics[width=\textwidth]{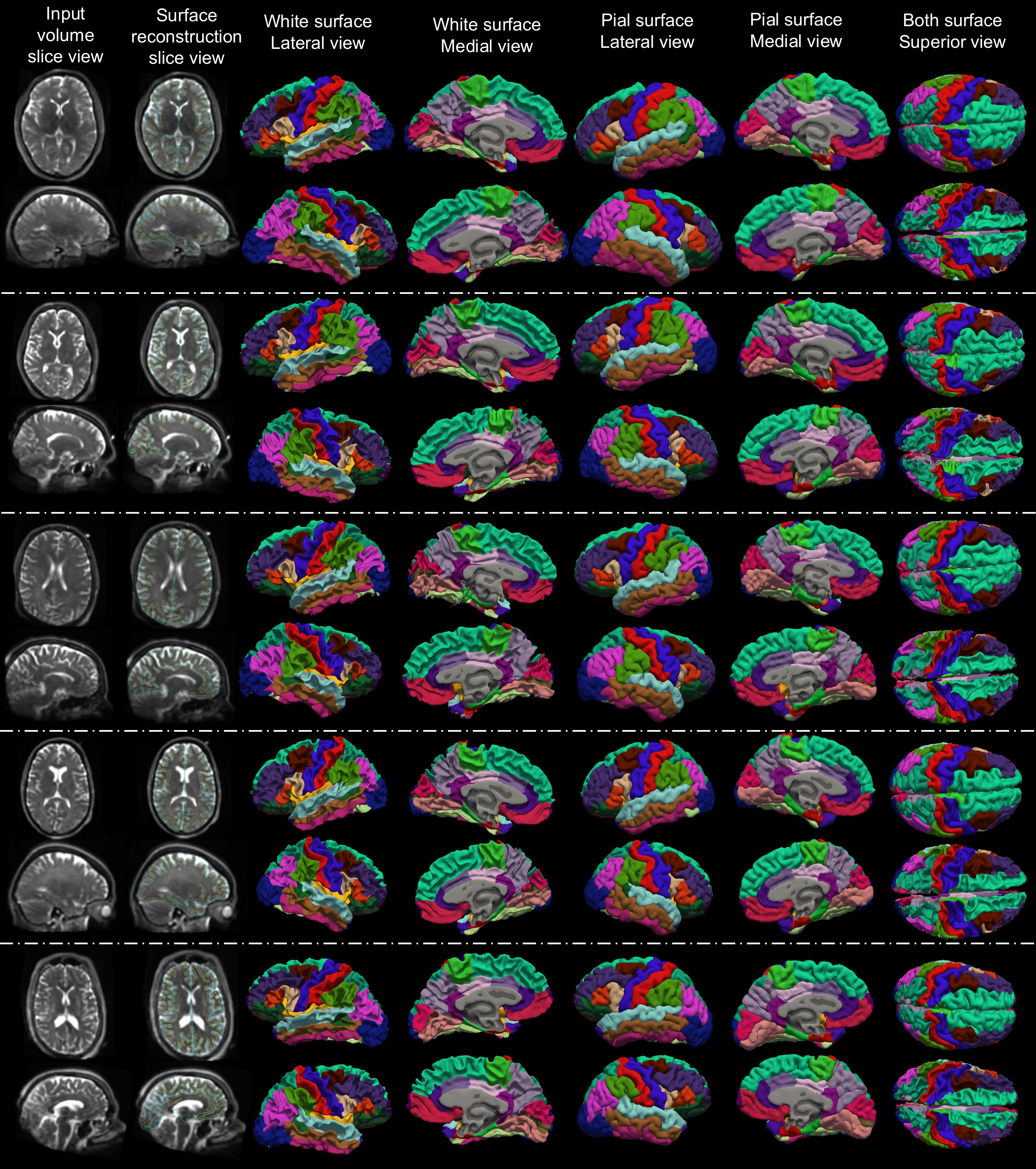}
        \end{figure*}

\end{document}